\documentclass[twoside,11pt]{article}
\usepackage{gensymb}
\usepackage{jmlr2e}

\usepackage{lastpage}
\jmlrheading{22}{2021}{1-\pageref{LastPage}}{1/21; Revised
7/21}{9/21}{21-0019}{Valerio Biscione and Jeffrey S. Bowers}

\ShortHeadings{CNNs Can Learn To Be Invariant To Translation}{Biscione and Bowers}
\firstpageno{1}

\begin{document}

\title{Convolutional Neural Networks Are Not Invariant to Translation, but They Can Learn to Be}

\author{\name Valerio\ Biscione \email valerio.biscione@bristol.ac.uk \\
       \addr Department of Psychology\\
       University of Bristol\\
       Bristol BS8 1TL, United Kingdom
       \AND
       \name Jeffrey\ S.\ Bowers \email j.bowers@bristol.ac.uk \\
       \addr Department of Psychology\\
       University of Bristol\\
       Bristol BS8 1TL, United Kingdom}

\editor{Amos Storkey}
\maketitle

\begin{abstract}
When seeing a new object, humans can immediately recognize it across different retinal locations: the internal object representation is invariant to translation. It is commonly believed that Convolutional Neural Networks (CNNs) are architecturally invariant to translation thanks to the convolution and/or pooling operations they are endowed with. In fact, several studies have found that these networks systematically fail to recognise new objects on untrained locations. In this work, we test a wide variety of CNNs architectures showing how, apart from DenseNet-121, none of the models tested was architecturally invariant to translation. Nevertheless, all of them could learn to be invariant to translation. We show how this can be achieved by pretraining on ImageNet, and it is sometimes possible with much simpler data sets when all the items are fully translated across the input canvas.  At the same time, this invariance can be disrupted by further training due to catastrophic forgetting/interference. These experiments show how pretraining a network on an environment with the right `latent' characteristics (a more naturalistic environment) can result in the network learning deep perceptual rules which would dramatically improve subsequent generalization.
\end{abstract}


\section{Introduction} \label{Intro}

Human have no difficulty in recognizing objects across a wide range of retinal locations.  Taking inspiration from biological models, artificial neural networks have been endowed with convolution and pooling operations \citep{LeCun1998, LeCun1990} in order to support this translation invariance.  It is often claimed that Convolutional Neural Networks (CNNs) are robust to translation thanks to their architecture.  For example: 
    \begin{quotation}
    ``[CNNs] have an architecture hard-wired for some translation-invariance while they rely heavily on learning through extensive data or data augmentation for invariance to other transformations''
    \citep{Han2020b} \end{quotation} 
    \begin{quotation}``Most deep learning networks make heavy use of a technique called convolution \citep{LeCun1989}, which constrains the neural connections in the network such that they innately capture a property known as transalational invariance. This is essentially the idea that an object can slide around an image while maintaining its identity; a circle in the top left can be presumed, even absent direct experience) to be the same as a circle in the bottom right.'' \citep{DeepLearningMarcus2018}\end{quotation}

See also \citet{LeCun1995, GensDomingosSymmetry, Marcos2016, Fukushima1980} for similar claims. 

While it is difficult to overstate the importance of convolution and pooling operations in deep learning, their ability to support translation invariance has been overestimated. In fact, multiple studies have reported highly limited translation invariance in CNNs \citep{Kauderer-Abrams2017a, Gong2014, Azulay2019, Chen2017, Blything2021}. For example, \cite{Blything2021} observed that the CNN VGG16 performed at chance levels at identifying objects when they were translated more than their own width. 



Two common misunderstanding have contributed to view that CNNs are architecturally invariant to translation. The first reflects a confusion between invariance and equivariance. The convolution operation is translationally equivariant, meaning that a transformation applied to the input is transferred to the output \citep{Lenc2019}. This transformation is inconsistent with invariance (where different inputs map onto the same output), and accordingly, the convolutions in CNNs do not insure invariance. Moreover, perfect equivariance can be lost in the convolutional layers \citep{Azulay2019, Zhang2019} through subsequent sub-sampling (implemented with pooling and striding operations, commonly used in almost any CNN). Therefore, overall, most modern CNNs are neither architecturally invariant nor perfectly equivariant to translation.


The other reason for overestimating the extent that CNNs are invariant to translation resides in the failure to distinguish between trained and `online' translation invariance (see \citealt{Bowers2016}). Trained invariance refers to the ability to correctly classify unseen instances of a class at a trained location.  For instance, a network trained to identify instances of dogs across the whole visual field will be able to identify a new image of a dog across multiple locations. This training is in fact common due to data-augmentation: jittering the training samples so that the network is trained on items across different locations \citep{Kauderer-Abrams2017a, Furukawa2017}. More relevant is the concept of online translation invariance: learning to identify an object at one location immediately affords the capacity to identify that object at multiple other locations, consistent with human performance \citep{BiedermannCooper1991, CooperBiedermann1992, Ellis1989, Fiser2001, Blything2021}. In this work we show that it is possible to train a model to support \textit{online} translation invariance. `Training' \textit{online} translation invariance, in which a network is trained to exhibit online translation invariance, should not be confused with `trained translation invariance' in which images can be identified across the canvas after training exemplar images across many locations.  It should also not be confused with `architectural' online invariance in which additional mechanisms are built into CNNs in order to support the invariance.

Online translation invariance is generally measured by training a network on images placed on a certain location (generally the center of a canvas), and then testing with the same images placed on untrained location. In many reports CNNs performed at chance level on untrained locations \citep{Kauderer-Abrams2017a, Gong2014, Azulay2019, Chen2017, Blything2021}. A few recent studies have attempted to address the lack of translation invariance by modifying the architectures of CNNs: \citet{Blything2021} shows that using Global Average Pooling on a VGG16 provides perfect invariance; \cite{SemihKayhan2020} and \cite{Zhang2019} introduced two additional minor modifications to CNNs architectures that also improved online translation invariance (discussed in more details in Section \ref{networks}), but they both only tested on very limited displacement. In this work we test their techniques with our experimental setup consisting or large displacements.


Two recent findings are in striking contrast with the previous literature: \citet{Han2020b} and \citet{Blything2021} both found that standard CNNs pretrained on one data set, without any architectural modifications, support online translation invariance for objects taken from another data set. By directly comparing networks pretrained on ImageNet with corresponding untrained networks, \citet{Blything2021} confirmed that it was the pretraining that led to robust online invariance, consistent with the online invariance observed with humans who are also "pretrained" on the natural world prior to testing them in the laboratory \citep{BiedermannCooper1991, CooperBiedermann1992, Ellis1989, Fiser2001, Blything2021}. See also \citet{BlythingCommentary2021}.

\section{Current Work}
In this work we focus on online translation invariance in a variety of classic CNNs, and some recent CNNs with architectural modifications designed to support online invariance. We show that, even though in most cases CNNs are not `architecturally' invariant, they can `learn' to be invariant to translation by extracting latent features of their visual environments (the data set). 


Why is this important? First, it is important to know how CNNs work, and there is currently confusion about how and when CNNs support translation invariance.  Second, a network that learns deep characteristic of its visual environment such as being invariant to translation should accelerate subsequent training, and accordingly, it is important to understand the conditions that foster invariance. In addition, because CNNs have been recently suggested as a model for the human brain \citep{Richards2019, Ma2020, Kriegeskorte, Zhuang2020}, it is important to understand if and how they can learn fundamental perceptual properties of human vision, of which invariance to translation is one \citep{Blything2021, Bowers2016, Koffka2013}.

The experiments are organized as follows. In Experiment 1 we tested several networks for architectural invariance, and whether pretraining on ImageNet actually improves performance when objects are trained at one location and test at novel locations. In Experiment 2 we tested whether fully-translated images taken from simpler data sets are sufficient to acquire translation invariance. In Experiment 3 and 4 we carried out additional investigations into the conditions in which online translation invariance is acquired. For example, we tested whether translation invariance could be acquired by training on the whole canvas \textit{without} translating the items in the canvas, whether translating objects over the majority of the canvas affords invariance on the untrained area, and whether fully translating a subset of classes in a data set results in online translation invariance for all classes.

\subsection{Data sets} \label{data sets}
We used a variety of data sets spanning a high range of complexity. For the experiments in Section \ref{ImageNet} and \ref{Beyond ImageNet} we used 5 data sets all consisting of 10 classes:
ShapeNet (a data set of 3D object which we rendered on a black background, more details in Section \ref{appShapeNet}), from \citet{Chang2015shapenet}; FashionMNIST (FMNIST), a MNIST-like data set of clothings \citep{fashionMNIST},  Kuzushiji MNIST (KMNIST), a MNIST-like data set of Japanese characters \citep{KMNIST}; MNIST, from \citet{LeCun1998}; and a subset of 10 images from the Leek data set used in \citet{Leek2016} and \citet{Blything2021}, in which each class consist of a single image of a 3D object (we call this data set LEEK10). Representative examples (and, for Leek10 the entire data set) are shown in Figure \ref{fig:Figure1}, top-left.  For the experiments in Sections \ref{WholeCanvas} and \ref{Generalizing Translation} we used the letter subset of the EMNIST data set \citep{EMNIST}, containing 26 classes (one for each letter), consisting of hand-written characters. We chose EMNIST as the design of these experiments required a greater number of classes.

For each of these data sets, we generated two versions: a \textbf{one-location} data set and a \textbf{fully-translated} data set. The one-location data set was generated by resizing each sample from each data set to $50 \times 50$ pixels and placing it on the leftmost-centered location on a black canvas of $224 \times 224$ pixels. Therefore no translation was used for this data set. In the fully-translated data set, each sample was similarly resized to $50 \times 50$ and was randomly placed in the canvas. The original sample was always completely visible. We did not apply any further data-augmentation.

\subsection{Networks} \label{networks}
We used 5 commonly used networks: ResNet-50 \citep{He2016}, VGG16, with and without Batch Normalization (VGG16 and VGG16bn) \citep{VGG16}, AlexNet \citep{NIPS2012_c399862d}, and DenseNet-121 \citep{Huang_2017_CVPR}, covering a wide variety of Convolutional feed-forward architectures. For experiments in Section \ref{ImageNet} we also used two networks that have been proposed as a way to increase robustness to translation in CNNs. \cite{Zhang2019} interposed an Anti-Aliasing filter between the max-pooling operation and related subsampling (the stride operation usually performed together with max-pooling) in order to better preserve shift-equivariance. This has been shown to provide a good degree of translation invariance on small diagonal shifts (16 pixels in a 32x32 canvas), but it is not clear whether it would provide any improvement for greater displacements. We applied the suggested Anti-Aliased Max Pooling to VGG16bn (we refer to this network as VGG16bn Anti-Aliased). \cite{SemihKayhan2020} suggested changing the zero padding size from 1 to 2, which results in applying each value in the filter on all values of the input map (Fully-Convolutional operation). The authors showed that this approach prevented the network from learning shortcuts related to the absolute spatial location of the stimulus in a canvas.  This improved the amount of invariance to translation on a small canvas (32x32) but it is again unclear whether the same benefits could be obtained on a bigger canvas. We tested a VGG16bn network with Fully-Convolutional operation (VGG16bn F-Conv).

\section{Experiment 1: Untrained vs Pretrained on ImageNet} \label{ImageNet}
In this experiment we assessed architectural online invariance of several networks when trained on a one-location data set and tested on the same, fully-translated data set. We also tested whether pretraining on ImageNet is enough to acquire translation invariance. 

The 7 CNN models, either untrained and pretrained on ImageNet (described in Section \ref{networks}), were (re)trained on a one-location data set and tested on the same data set, fully-translated. The items used in the testing data set were the same as those used for training in the one-location training data set. That is, we did not use held-out samples, as we wanted to test on invariance to translation, and not the generalization ability that different networks may possess in differing degrees. The networks that were pretrained on ImageNet would therefore use the learned parameters as initialization for the new training with the one-location data set. This is commonly referred as fine-tuning, \citep{Girshick2014}. We did not freeze any layer when fine-tuning. The untrained network was initialized with Kaiming Initialization \citep{He2016}.
\begin{figure}[!ht]
\centering
  \includegraphics[width=1\linewidth]{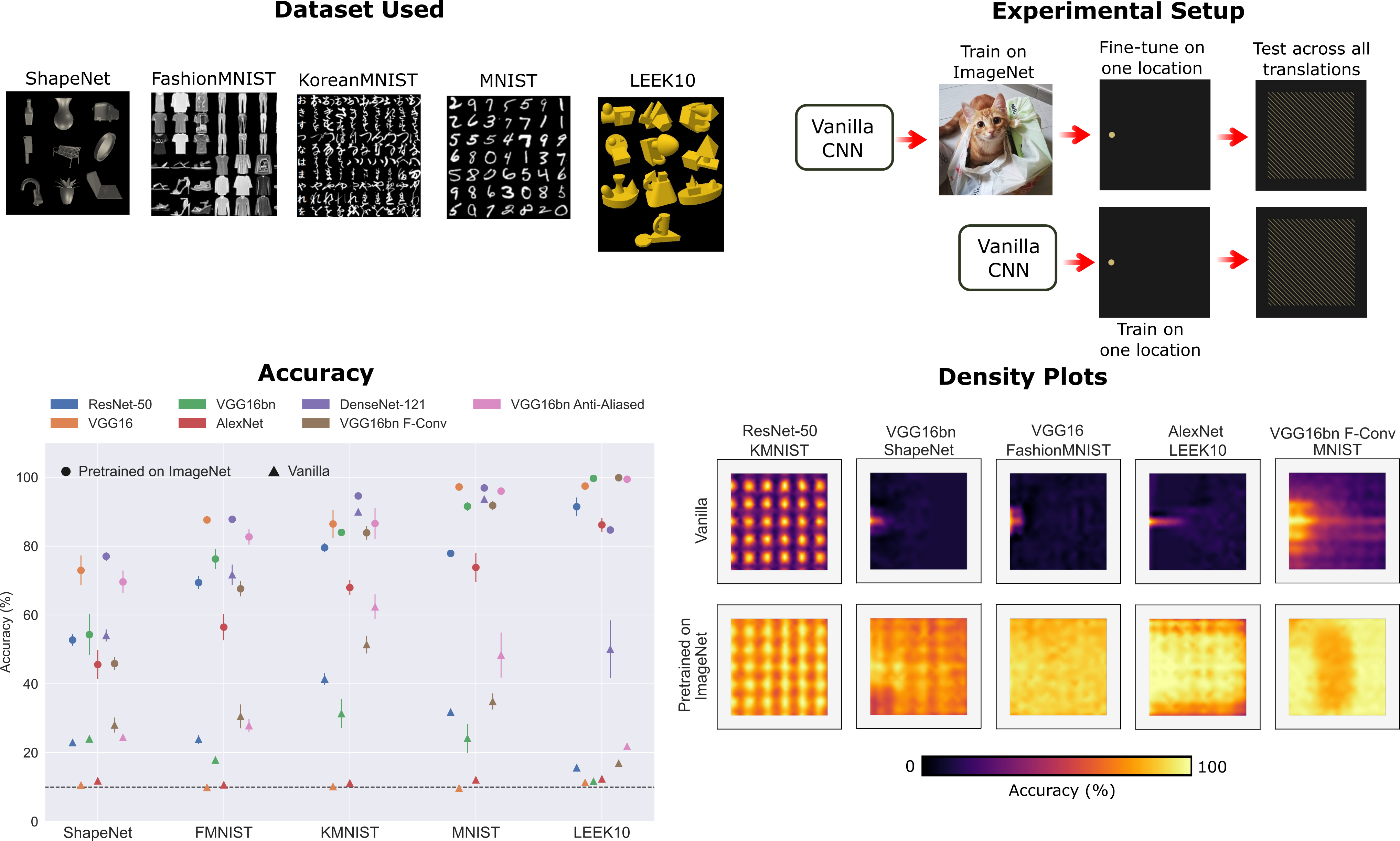}
\caption{\textbf{Top-left}. Samples of the 5 data sets used in this experiment. LEEK10 is shown in its entirety. \textbf{Top-right}. Representation of the general experimental design used here and in following experiments. \textbf{Bottom-left}. Accuracy results of networks trained on one-location data sets when tested on the same, fully-translate data sets. Circles refer to network pretrained on ImageNet, whereas triangles refers to untrained networks. Each point is averaged across the 3 runs. Dashed lines indicates chance level. All the networks reached at least 95\% accuracy on the one-location data set, so any performance lower than that indicates an imperfect invariance to translation. Networks pretrained on ImageNet were always resulting in a higher accuracy on the fully-translated data sets. \textbf{Bottom-right}. The heatmaps show the accuracy for some networks and data set across the whole canvas. These are only few representative examples. All the Density Plots are shown in Section \ref{appDensity}.}
  \label{fig:Figure1}
  \end{figure}

We used Adam optimizer with a fixed learning rate of 0,001, and trained until convergence. All networks, both pretrained on ImageNet and untrained, reached at least 95\% of accuracy on the one-location data set. We repeated each training across 3 seeds. The experimental design is shown in Figure \ref{fig:Figure1}, top-right.

Since all networks were performing at high accuracy on the one-location data set, a network possessing a high degree of online translation invariance would also achieve a high accuracy on the same fully-translated data set (recognising trained objects at unseen locations).  The  average accuracy results are shown in Figure \ref{fig:Figure1}, bottom-left. 

We can have a more fine-grained understanding of each network mechanism by computing an ``accuracy density plot": we tested the network on items centered across a grid of $19 \times 19$ points equally distributed upon the canvas. By interpolating the results across the tested area we obtained a heatmap like those shown in Figure \ref{fig:Figure1}, bottom-right. Here we only present some interesting patterns found in the data set, with the full set of Density Plots shown in Section \ref{appDensity}.

In all cases, networks pretrained on ImageNet showed higher accuracy on unseen locations than untrained networks, confirming and extending the findings from \citep{Blything2021}. This is all the more notable when considering the major differences between the features in ImageNet and the data sets used here. There were however some significant differences amongst networks and data sets.

It was surprising to find that an untrained ResNet-50 did not support online translation invariance even though it possesses a Global Average Pooling (GAP) layer.  As noted above a GAP layer is common architectural modification to CNNs that has been used to obtain online translation invariance in other untrained networks such as VGG11bn \citep{Blything2021}. As a sanity check, we confirmed that an untrained VGG16bn and VGG16 (two of the networks used in this experiment), endowed with a GAP layer after the convolutional blocks, possessed indeed a high degree of translation invariance (~98\%).  It is unclear why the GAP layer provides translation invariance for VGG11bn, VGG16bn, and VGG16, but not for ResNet-50. One hint could be provided by the density plot, which presents a checkerboard pattern generally attributed to the striding operation \citep{Zhang2019}.

From the experiments it is also possible to infer the role of Batch Normalization. AlexNet and VGG16 were the only two networks without Batch Normalization, and they consistently performed at chance across all data sets in their untrained version. Comparing VGG16 and VGG16bn (green and orange triangles in Figure \ref{fig:Figure1}) shows that BatchNorm provides a small but consistent advantage in recognizing objects in novel locations with untrained networks. Surprisingly, in the ImageNet-pretrained condition, VGG16bn performed \textit{worse} than VGG16 when tested on novel locations, for all data sets but LEEK10. 
DenseNet-121 was the only network that consistently reached a high accuracy for most data sets even in its untrained version (although still lower than when pretrained on ImageNet), meaning that it does possess a certain degree of architectural translation invariance. It is unclear why this is the case but it may be due to the fact that each layer of this network can access information from the earlier layer and the input image. 

Finally, the two networks with architectural modifications designed to improve translation invariance (VGG16bn F-Conv and VGG16bn Anti-Aliased) do seem to achieve slightly better results than their not-modified counterpart (VGG16bn), especially in the KMNIST and MNIST data set. However, when pretrained on ImageNet, their accuracy is still greatly increased, indicating a much higher degree of online translation invariance obtained by pretraining on ImageNet. 

Regarding the differences amongst one-location data sets, as data set complexity increased, networks pretrained on ImageNet showed a decreasing performance on the fully-translated data set, (e.g. networks fine-tuned on ShapeNet exhibited a much worse translation invariance than networks fine-tuned on the simpler MNIST variants or LEEK10). This point will be investigated in more details in the next Section. 

As for the density plots, for most untrained networks high accuracy was achieved only at trained location (the leftmost-centered one), and performance degraded quickly after short displacement. Most networks pretrained on ImageNet had a significantly more uniform distribution of error. However different networks presented slightly different patterns, which also seemed to depend on the data set used (see Section \ref{appDensity}).

\subsection{Experiment 2: Acquiring Translation Invariance with Translated data sets} \label{Beyond ImageNet}
We hypothesised that the network pretrained on ImageNet learned translation invariance because of the data augmentation performed during pretraining (``Random Crop") which resulted in the network being trained on translated versions of the same objects across the visual field. That is, as a side effect of the data-augmentation, the network was trained on a fully-translated data set. If this is correct, we could hypothetically make a network learn to be online-invariant to translation with any simple data sets in which the items occur all across the canvas. We stress that this means that the network would learn to be invariant to \emph{novel} objects from \emph{novel} classes. There are obviously other features that a network can extract from ImageNet that are not related to image translation, but here we will focus on translation only.

\begin{figure}[!ht]
\centering
  \includegraphics[width=1\linewidth]{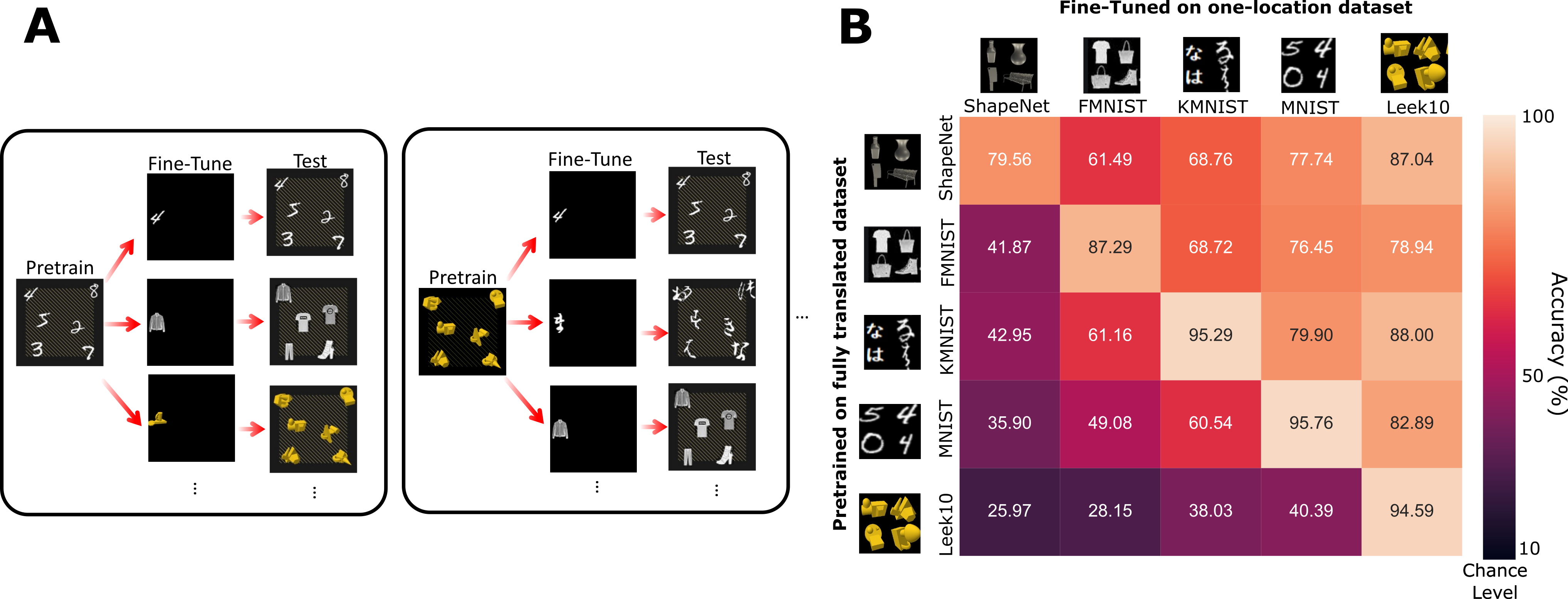}
\caption{\textbf{A}. Experimental design for Experiment 2. Each network is trained on a fully-translated data set, then fine-tuned on each one-location data sets, and finally tested on the fully-translated version of the fine-tuned data set to assess the degree of translation invariance. Multiple samples are shown in the same canvas just for illustrative purpose - in practice, only one sample was present in the canvas at any given time. \textbf{B}. Results from Experiment 2, averaged across networks and 3 repetitions. Most networks performed far better than chance when pretrained on a fully translated data set, but there is a clear effect of the pretraining data set. Each network presented difference in absolute accuracy, but all of them presented the same pattern in which a better performance was obtained when the pretraining data set was more complex than the fine-tuning one. Detailed results for each network and data set are shown in Section \ref{AppBeyond}}
  \label{fig:Figure2}
   \end{figure}

In order to test this hypothesis, instead of using networks pretrained on ImageNet, we pretrained 4 networks (ResNet-50, VGG16, VGG16bn, and AlexNet) on the fully-translated data set in Section \ref{data sets}.
Similarly to the previous experiment, we then fine-tuned the networks on each one-location data set, and tested it on the same items, fully-translated (see Figure \ref{fig:Figure2}A for an illustration of the experimental setup). Each network is trained until convergence, which in almost all cases corresponded to an accuracy of at least 95\% for both the pretraining phase and the one-location fine-tuning. Again, if the networks have learned to be invariant to translation, they would be able to recognise objects trained at one location everywhere on the canvas, without the need to be trained on every location. The resulting accuracy, averaged across the whole canvas, networks, and 3 repetitions, are shown in Figure \ref{fig:Figure2}B.
 In most cases, the networks were far above chance when items that were trained one location were now displayed across the whole canvas (compare this results with the bottom-left of Figure \ref{fig:Figure1} where we show the performance of a untrained network trained on one-location and tested on the whole canvas). 
 
 Two results were striking. Firstly, some combinations of pretrained and fine-tuned data sets resulted in low accuracy. We observed a pattern in which networks pretrained with a ``complex'' fully-translated data set obtained high performances when fine-tuned on a one-location ``simple'' data sets, but the opposite was not true (for example, pretraining on LEEK10 resulted in poor performance when fine-tuned on any other data set)\footnote{For FashionMNIST, KMNIST and MNIST we based our judgment of complexity on the benchamrk in \url{https://paperswithcode.com/task/image-classification}. We deemed Leek10 as the easiest to classify due to the lack of intra-class variability, as each class is composed of just one image, and ShapeNet as the most difficult due to its high intra-class variability, and the complexity of the 3D objects.}. Also, pretraining on MNIST supported translation invariance for the simpler LEEK10, and in a lesser degree for KMNIST, but not for the more difficult ShapeNet and FMNIST.
 
 The second important result is that when pretrained and fine-tuned \textit{on the same data set} the accuracy would actually sometime go down when tested on the fully-translated data set again (this happened for ShapeNet and FMNIST). To clarify this point: a network pretrained on fully-translated ShapeNet (reaching therefore an accuracy of at least 95\%), and then fine-tuned on a one-location ShapeNet data set (still obtaining an high accuracy on training data), would now suffer a drop in accuracy to 79.56\% when retested on the fully-translated ShapeNet. This surprising result seem to be due to an acquisition of a bias by the fine-tuned network, in which the network learnt to ignore areas of the canvas where items were not present during the fine-tuned training session.
 
Both of these findings could be explained by the phenomenon of catastrophic interference  \citep{Furlanello2016, McCloskey1989} in which training on new tasks degrades previously acquired capabilities. If this is correct, it might mean that all networks tested above had indeed acquired online-invariance to translation, but in some cases this was subsequently lost due to fine-tuning, which would `overwrite' the weights responsible for translation-invariance. Fine-tuning on more difficult data sets would result in a larger number of weights needed to be adjusted, which would in turn result in lower performance when tested on fully-translated objects. Conversely, networks pretrained on complex features may have required little weight adjustments and would more likely preserve the weights coding the invariance to translation. 
Measuring translation invariance through performance on novel locations after fine-tuning on a one-location data set could underestimate the degree of translation invariance acquired when the network is pretrained on a fully-translated data set. Network could acquire online-invariance to translation \textit{with any translated data set}, regardless of its complexity, and this ability could be retained with techniques that prevent catastrophic forgetting (\citealt{Beaulieu2020, OML2019,  Li2016}). 
To test this, we adopted an alternative measure of translation invariance. 

  \subsection{Measuring Translation Invariance Through Similarity of Translated Representations} \label{representation}

In order to test the hypothesis that catastrophic interference played a role, we computed an Invariance Metric, a measure of activation similarity between translated versions of the same sample, normalized by the similarity across different samples (details in Section \ref{metric}). The Invariance Metric was computed between the input image at the leftmost-centered location and the corresponding image at different horizontal displacement at the penultimate layer. This metric indicates the degree to which translated objects share the same internal representation.


\begin{figure}[!b]
\centering
\includegraphics[width=0.65\linewidth]{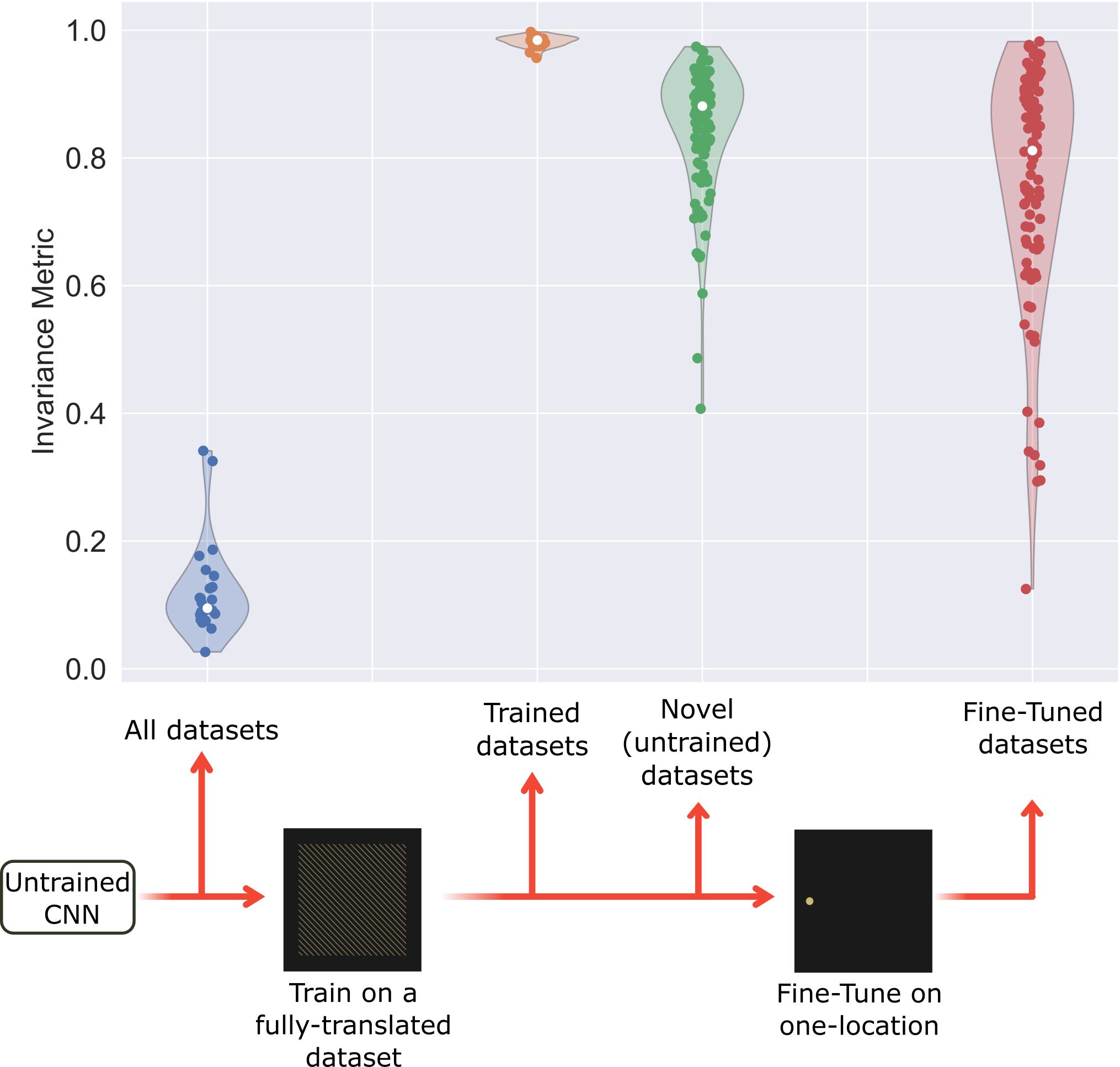}
\caption{Invariance Metric at different stages of the experimental setup for experiments in Section \ref{Beyond ImageNet}.  The bottom part of the figure indicates the different stages at which the analysis is performed. The Invariance Metric indicates the degree of translation invariance possessed by the networks. Each point represent the result for a network and a data set. After training on a fully-translated data set, the networks show translation invariance (green circles), but this gets disrupted after fine-tuning (red circles). Detailed results for each horizontal displacements are shown in Section \ref{appHorizontal} and \ref{appInvarianceAll}}
  \label{Figure3}
   \end{figure}

The degree of catastrophic forgetting can be determined by computing the average Invariance Metric across several horizontal displacements, at different experimental stages (see Figure \ref{Figure3}).  With an untrained network, the Invariance Metric was in almost all cases $< 0.2$, indicating low similarity across translated versions of the same object. After trained on fully-translated version of a data set, we computed the Invariance Metric across that same data set, which was near $1$ (this would be a measure of `training' invariance). The same networks was then tested on  \textit{novel} data sets, that is data sets that are different from the one the each network has been trained on. This resulted in a high Invariance Metric (0.83$\pm$0.1) which corresponds to a high degree of online translation invariance. Therefore, the network was able to acquire translation invariance on \textit{novel} data sets by being trained on a fully-translated data set. However, following fine-tuning a network on a one-location data set, we observed \textit{lower} translation invariance on average with our Invariance Metric, which can be only explained by catastrophic interference happening during the fine-tuning stages (more detailed results in Section \ref{appHorizontal} and \ref{appInvarianceAll}).

\section{Further Experiments on Translation Invariance}
In the following Sections we describe a series of experiment aimed to understand the limits of translation invariance across several experimental conditions. We focused our exploration on the VGG16 network, and we used the EMNIST data set. VGG16 was the network showing the most extreme improvement in accuracy when pretrained on a fully-translated network (either ImageNet or any of the other translated data sets), and thus was an ideal candidate for further tests.  The EMNIST data set is particularly useful as the high number of classes (26) makes it ideal for conducting experiment with split classes (as detailed below). 

We firstly confirmed that the results obtained in Section \ref{ImageNet} and \ref{Beyond ImageNet} replicates with EMNIST: VGG16 pretrained on ImageNet and Fine-Tuned on EMNIST at one location obtained a high level of translation invariance (84$\pm$3.12\%  accuracy) compared to an untrained network (around 5$\pm$0.76\%, close to chance level which is 3.8\%). We also verified that VGG16 pretrained on EMNIST and fully-translated and fine-tuned on MNIST at one location would show a high degree of online translation invariance when tested on translated MNIST data set (96.8$\pm$2.81\%), which indicated a low degree of catastrophic interference. All these tests were performed across 3 seeds. 

\begin{figure}[!b]
\centering
  \includegraphics[width=1\linewidth]{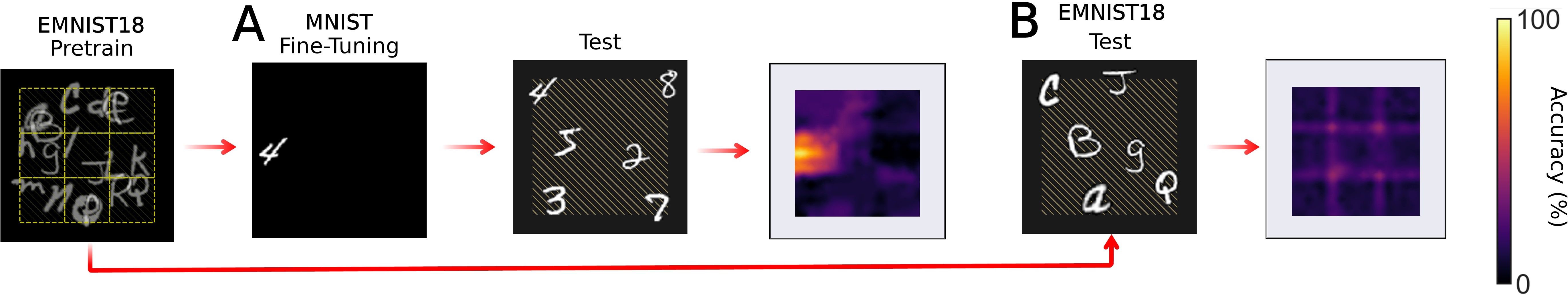}
\caption{Experimental design for Experiment 3 and results. We pretrained on a data set in which different letters categories were shown in different area of the canvas. In this way the network is trained on the full-canvas, but not on a fully-translated data set. \textbf{A}. After fine-tuning on MNIST, the network did not generalize on untrained locations. \textbf{B}. When tested on the same data set (EMNIST18), but without using classes segregation (each letter presented at all locations), letters were only recognised when presented on the area they were trained on, so mean accuracy was low. In both cases, the heatmaps are averaged across all classes.}
\label{Figure4}
   \end{figure}
\subsection{Experiment 3: Training on the Whole Canvas Is Not Enough} \label{WholeCanvas}
It is conceivable that in the previous experiments, and in similarly designed experiments in the literature \citep{Kauderer-Abrams2017a, Gong2014, Chen2017, Blything2021}, untrained networks failed to show invariance to translation because they were tested on locations where they had not seen any items. In which case, pretrained networks succeeded not because they had acquired the deep property of translation invariance from the visual environment, but simply because they had been trained on the whole canvas. To test this hypothesis we separated the canvas in 9 equilateral areas ($58 \times 58$ pixels), and within each area, only 2 of the classes were presented. The items were randomly centered anywhere within their area (in such a way that part of the object could sometime slightly overlap another area, but the objects were never cropped). Therefore the objects were subjected to highly limited translation.  Once trained with this setup, we  determined the degree of translation invariance in two ways: fine-tuned the network on a one-location MNIST data set and tested the same data set fully-translated (like in Section \ref{ImageNet} and \ref{Beyond ImageNet}). We used the EMNIST data set with the first 18 categories (EMNIST18) for pretraining. Results are shown in Figure \ref{Figure4}A. We also tested the pretrained network on a fully-translated version of EMNIST18, that is, without class segregation, and without fine-tuning (Figure \ref{Figure4}B). Even though the network was trained on items everywhere on the canvas, it did not acquire the ability to generalize on unseen locations with neither EMNIST18 nor newly trained objects (MNIST). This is a strong demonstration that the network needs to be trained on an environment where objects are translated across the whole canvas in order to learn to be invariant to translation, and that simply training on the whole canvas is not enough.

\subsection{Experiment 4: Training on Limited Translations Is Not Enough} \label{Generalizing Translation}
In the previous experiments we assessed online translation invariance after training networks with objects at one location. We showed that a network would display invariance only if it had been pretrained on a fully-translated data set. Here we explore alternative visual environments that are somehow `limited' in the way translation invariance was present across the training data set. We then tested whether translation invariance was nevertheless acquired by measuring the performance on the fully-translated data set. 
We divided the canvas in 4 quadrants and treated the upper-right quadrant in a special way in the following three conditions. In each case, we used letters from the EMNIST data set.

\begin{figure}[!b]
\centering
  \includegraphics[width=1\linewidth]{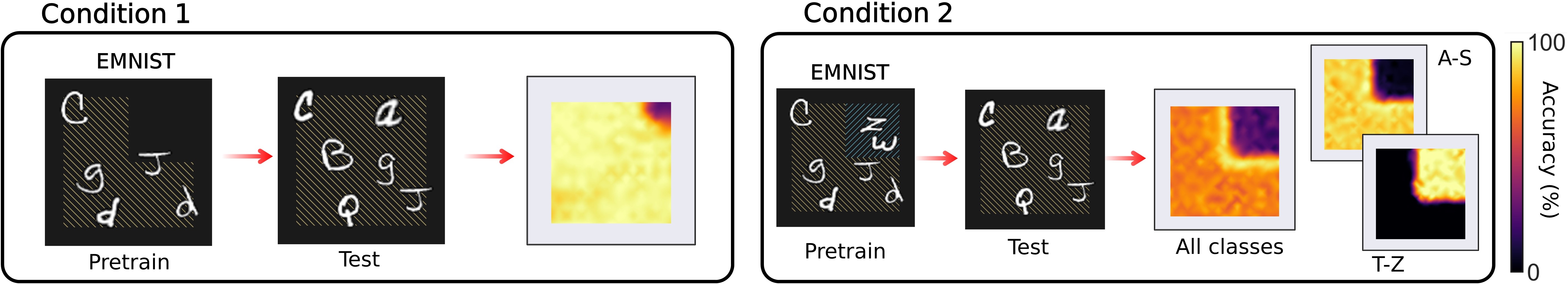}
\caption{Experimental design and results of condition 1 and 2 in Experiment 4. In both cases, we trained and tested on the same EMNIST data set. In Condition 1, the translation was limited to three quarter of the canvas. The heatmap shows a certain degree of generalization on untrained area, but not complete. On Condition 2, we limit translation but still train on the upper-right area, with limited classes. The network only accurately predicted classes on the area in which they were present during training.}
  \label{Figure5}
   \end{figure}

In \textbf{Condition 1} the upper-right quadrant was left empty, and all the other classes were trained on the rest on the canvas. When tested, the network was only partially able to generalize on the untrained area (Figure \ref{Figure5}, left).

In Conditions 2 and 3 we explored the possibility that training different classes on different areas would allow the network to recognize the classes outside their training area. Unlike the experiments in Section \ref{Beyond ImageNet}, in which new classes were trained at one location \textit{after} the fully-translated pretraining had occurred, here we trained the fully-translated classes and partially translated classes at the same time. 

In \textbf{Condition 2} we trained the letter T to Z of the EMNIST data sets on the upper-right quadrant, and A to S on the rest of the canvas (excluding the upper-right). When tested on a fully-translated version on the full A to S data set the network failed: the letters would be recognized only in regions they were trained (Figure \ref{Figure5}, right).

In \textbf{Condition 3} we slightly modified the previous experiment by training a subset of classes across the whole canvas, while limiting another subset to the upper-right quadrant.  We split the experiment in 4 sub-conditions, varying the number of fully-translated and partially-translated classes. In all sub-conditions the network did not achieve full translation invariance on classes that were presented only on the upper-right quadrant (more details in Section \ref{Cond3} and Figure \ref{FigureAppCondition3}).

The overall conclusion from these experiments is that, in order to acquire the property of being invariant to translation, a network needs to be pretrained on fully-translated objects, for all classes. Further details on these experiments, including the results when fine-tuned on a new data set, are provided in Section \ref{AppGeneralizing}.

\section{Discussion} \label{Discussion}
The aim of this work was twofold: first, we assessed whether commonly used CNN model possess architectural translation invariance, that is the ability to recognize objects on novel location after being trained on one location, as commonly claimed. Secondly, we assessed the degree to which CNN can learn online translation invariance through pretraining. We found that all networks but one, DenseNet-121, suffered from a significant and consistent drop in performance when trained on one location and tested on the full canvas. That is, most CNNs do not possess architectural invariance.  Furthermore, various architectural modifications of CNNs designed to increase online invariance only had limited success. Perhaps most surprisingly, Global Average Pooling that supports full translation invariance in some networks \citep{Blything2021} does not work in all cases: ResNet-50 presented a checkerboard pattern which overall resulted in poor performance. Similarly, Anti-Aliasing from \cite{Zhang2019} and Fully Convolutional operation with zero padding \cite{SemihKayhan2020} showed limited improvement in performance upon their unmodified counterparts, and only with some of the tested data sets. 

By contrast, we found that pretraining on a fully translated data set improved online translation invariance consistently. This was demonstrated most clearly through an Invariance Metric that measured the similarity of hidden unit activations in the penultimate layer of the various networks following translations of novel objects.  Critically, this invariance was achieved not only after training on large and complex data sets (e.g., ImageNet) but also on translated images taken from simple data sets (e.g., MNIST).  However, this online invariance was not always manifest in performance when we included additional training of images at one location (fine tuning) and then assessed performance at novel locations. Our analysis showed that online translation performance was sometimes impaired by a catastrophic intereference, and that a critical challenge for future work is to prevent new learning from impairing the trained invariances that are acquired in the hidden layers of the networks.

Our experiments show that translation invariance can be learned in CNNs that lack built-in architectural invariance, and raise the possibility that a whole range of perceptual capabilities can be learned rather than built in the architecture. This work also suggest that, in certain cases, the architecture may be less important than the visual world the network is trained on.

Neural networks performance is often compared to human performance \citep{Baker2018, HanYoon2019, Srivastava2019, Ma2020}. A fundamental feature of human perception is that it supports widespread generalization, including combinatorial generalization (e.g., identifying and understanding images composed of novel combination of known features).  Current CNNs are poor at generalizing to novel environments \citep{Geirhos2020}, especially when combinatorial generalization is required   \citep{VankovBowers2020, HUmmel2013, Montero2021TheGeneralisation}. Here we have shown that CNNs are able to extract latent principles of translation invariance from their visual world and to re-use them to identify novel stimuli with very different visual forms in untrained locations, which constitutes a weak form of combinatorial generalization. An interesting question is the extent to which stronger instances of generalization and other fundamental principles of perception (e.g., Gestalt principles of organization, \citealt{Koffka2013}) can be learned in standard CNNs trained on the appropriate data sets, and what sorts of generalization requires architectural innovations. Without the right training environment, it is not surprising that CNNs fail to capture the cognitive capacity of the human visual system \citep{FunkeBorowski2020}, and the only way to address this fundamental question is to train models under more realistic conditions.

Although training on more naturalistic data sets may lead to better and more human-like forms of generalization, it is also worth highlighting the contribution of our experiments to the wider field of Machine Learning. We showed that artificial hand-crafted environments, such as our fully-translated data sets, could be used to train networks to acquire a particular perceptual regularity in spite of the architecture not directly supporting that invariance. If the challenge of catastrophic interference can be overcome, this may prove to be a useful technique for learning in more complex environments: Instead of having a network learning all possible visual configurations (through data augmentation) of a given data set, the network can be pretrained on extremely simple data sets that embed the fundamental perceptual principles of the environment. Subsequently, when facing a more complex data set, the network would only need to learn the basic configuration of the new objects, and generalize to others through the learned perceptual invariances. 




\section{Conclusion}
We have shown that, even though most CNNs are not architecturally invariant to translation, they can learn to be by training on a data set that contains this regularity. We have also shown how this property is retained when fine-tuning on simpler data set, but lost for more complex ones, and this appears to reflect the role of catastrophic interference in constraining translation invariance rather than a failure to learn invariance from these simple data sets. More generally, we suggest that by training artificial networks with non-naturalistic data sets (e.g. data sets that do not contain regularities normally found in human visual environments), we are constraining their ability to learn deep fundamental rules of perception.


\acks{This project has received funding from the European Research Council (ERC) under the European Union’s Horizon 2020 research and innovation programme (grant agreement No 741134). We would like to thank Gaurav Malhotra for helpful comments and suggestions.}

\newpage

\appendix
\section{ShapeNet data set Generation} \label{appShapeNet}
We selected 10 classes out of the 55 included in ShapeNet. They were: faucet, bench, laptop, rifle, table, jar, knife, bottle, clock, bus. For each class, we randomly selected $250$ 3D objects. Each 3D object was rendered textureless on a uniform black background as a 224$\times$224$\times$3 image. To obtain variation in viewpoint, we placed a camera on a sphere at an inclination ranging from 30\degree to 110\degree (10\degree intervals) and an azimuth covering the whole sphere (36\degree intervals generating 90 different viewpoints per object). Amongst these, we selected 30 viewpoints for each object, totalling 6000 images per class.

\section{Untrained vs Pretrained on ImageNet: Density Plots} \label{appDensity}
\begin{figure}[!ht]
\centering
  \includegraphics[width=1\linewidth]{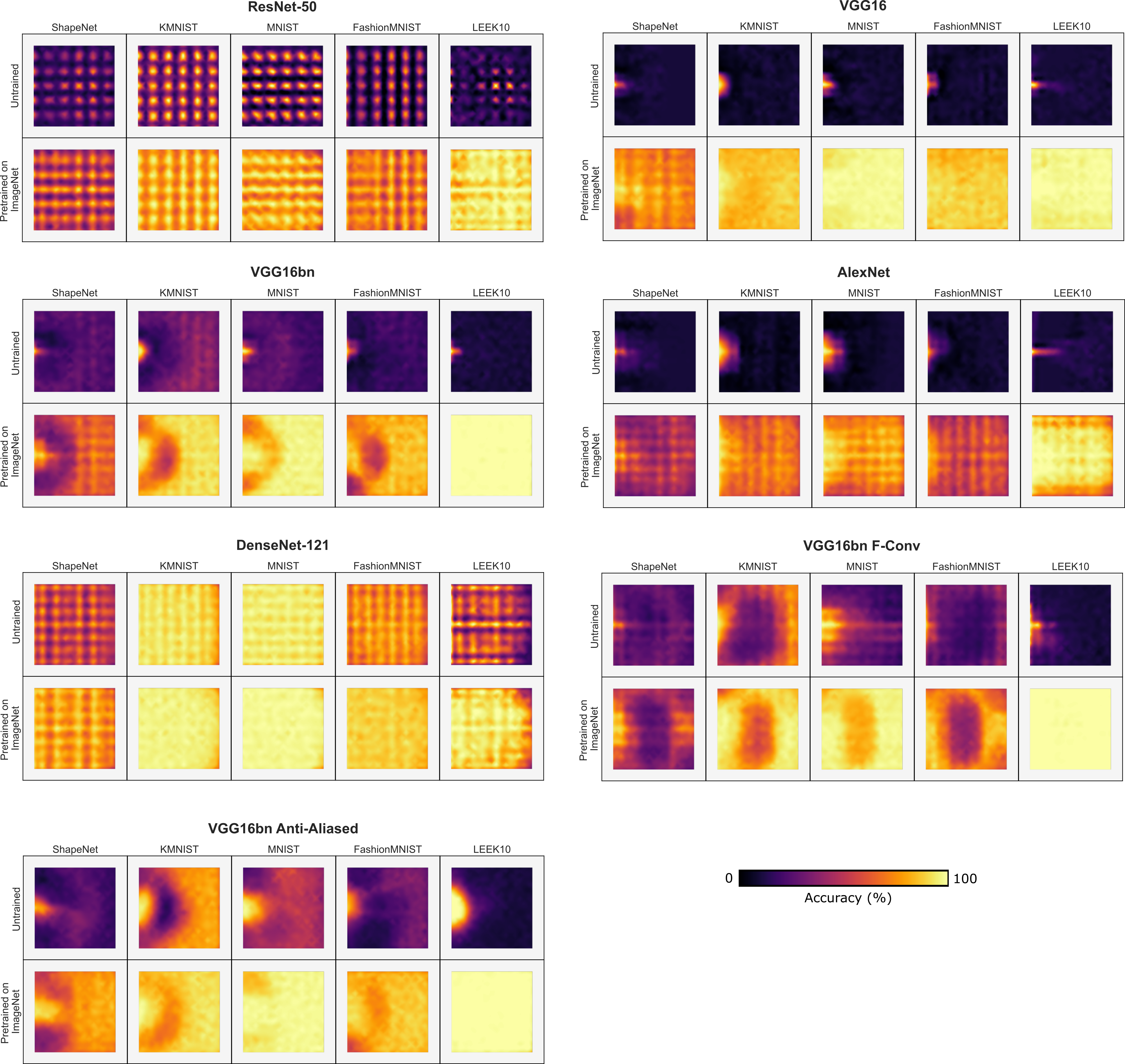}
\caption{Density Plot for each network and data set. Networks, untrained or pretrained on ImageNet, are trained/fine-tuned on one-location data sets specified in each column. They are then tested on the same data set, now fully-translated.}
  \label{FigureAppDensity}
   \end{figure}

In Section \ref{ImageNet} we only showed few examples of the Density Plots resulting from testing the performance of each network across the whole canvas. In Figure \ref{FigureAppDensity} we show the results for each networks and data sets, for untrained conditions and pretrained on ImageNet conditions. Each Density Plot is averaged across 3 repetitions. 
 
\section{Acquiring Translation InvarianceWwith Translated Data Sets: Individual Networks Results} \label{AppBeyond}
\begin{figure}[!ht]
\centering
  \includegraphics[width=1\linewidth]{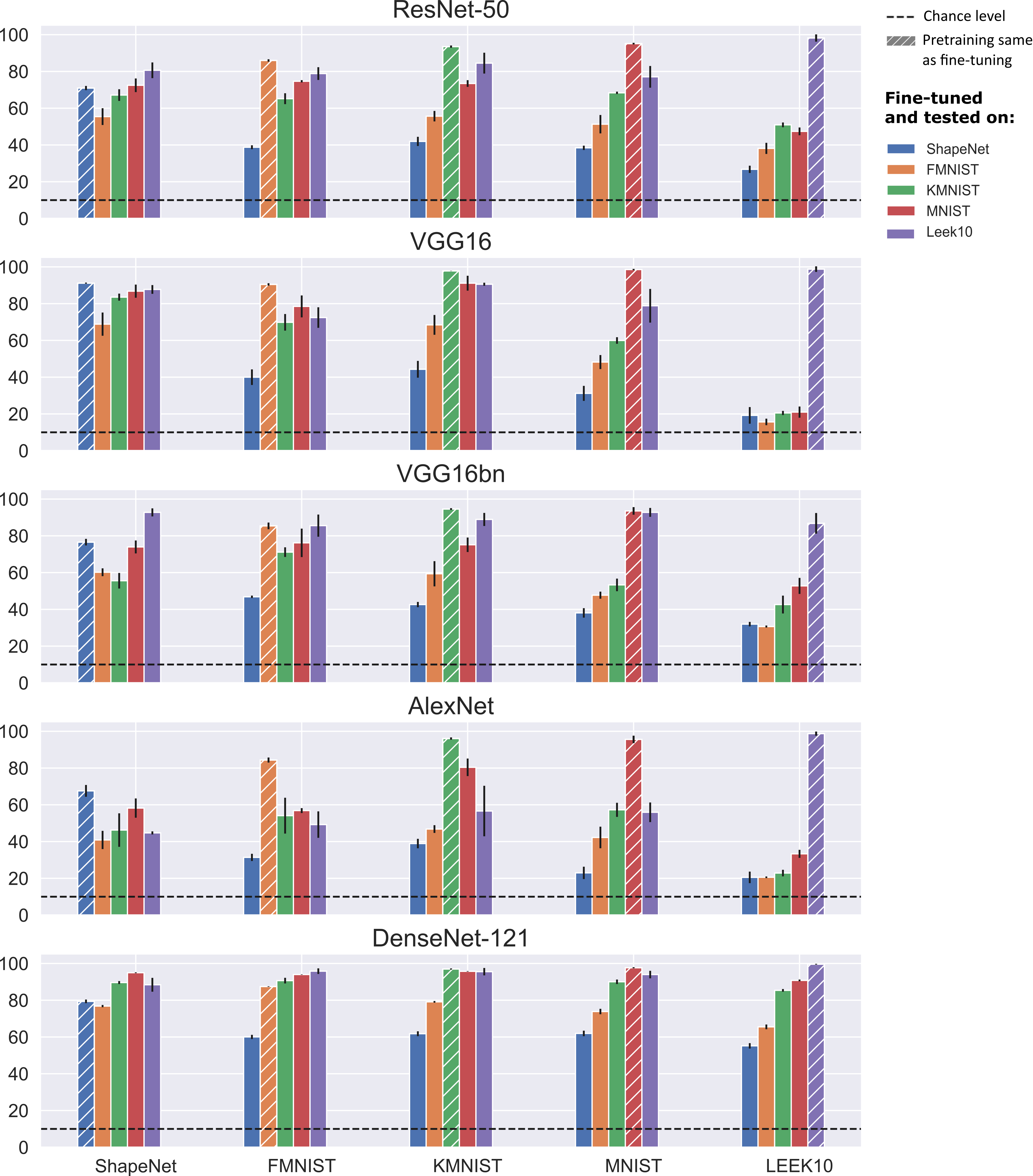}
\caption{Average results across each repetition for Experiment 1. Error lines indicate one standard deviation. Hatches indicate the condition in which fine-tuning had the same items as pretraining (corresponding to the diagonal in Figure \ref{Beyond ImageNet})}
  \label{FigureA9}
   \end{figure}

In Section \ref{Beyond ImageNet} we pretrained on the fully-translated version of each one of the the 5 data sets and for 4 networks: ResNet-50, VGG16, VGG16bn, and AlexNet, averaging across all networks for compactness. We present in Figure \ref{FigureA9} the full results for each individual network. Each network was pretrained on the data set indicated by the position in the horizontal axis' plot, fine-tuned on the data set indicated by the bar color, and tested on the fully-translated version of that same data set. We repeated each train/fine-tune/test procedure 3 times. Error bars indicate standard deviation across repetitions. 

\section{Invariance Metric} \label{metric}
In Section \ref{representation} we used the Invariance Metric, a measure based on cosine similarity to measure the similarity across translated version of the \textit{same} objects, by normalizing across representation of \textit{different} objects.

For each network's layer, we computed the cosine similarity across activations between input samples $s_i$ and $s_j$ as follows:

$$C^l(s_i, s_j) = \frac{d^l(s_i) \cdot d^l(s_j)}{\left \| d^l(s_i) \right \| \left \| d^l(s_j) \right \|},$$

where $d^l(\cdot)$ is the activation at layer $l$ given input $a$.

Let's consider a single item $o_r$ such as a specific image representing a letter from MNIST. We define $t(o_r, \theta)$ as a resize and translation operation which resizes $o_r$ to 50$\times$50 pixels, and place it within a 224$\times$224 black canvas, to a location determined by parameter $\theta$. 


Now consider a specific location parameter $\hat{\theta}= (x=25, y=112)$ that we use as a base-translation for performing the similarity comparison.
We defined $T(\theta)$ as the average similarity of representations between the objects base-view and a translation defined by parameter $\theta$ ($T$ for \textit{Translation Similarity}) across $R$ items $o = \{o_1, ..., o_R\}$:
$$T(\theta) = \frac{1}{R}\sum_r^R C^l(t_{\hat{\theta}}(o_r), t_\theta(o_r)).
$$

We further define $U(\theta)$ (for \textit{Uniformity}) as the between-objects invariance, that is the total average similarity between $N$ different randomly sampled items
$u = \{u_1, ..., u_N\}$ and $v = \{v_1, ... , v_N\}$:
$$U(\theta) = \frac{1}{N}\sum_n^N C^l(t_{\hat{\theta}}(u_n), t_{\theta}(v_n)).$$

We used $U(\theta)$ as a baseline across object translations: if a model has learnt translation invariance, $T(\theta)$ should be higher than $U(\theta)$. Therefore we defined the Invariance Metric as

$$ I(\theta) = \frac{T(\theta) - U(\theta)}{1-U(\theta)},$$

which corresponds to the invariance across translation $\theta$ that are only due to translation invariance (and not to being insensitive to different objects). A value of 1 corresponds to identical representation across translated versions of the same item; a value of 0 indicates that two translations of the same items are as different as two different items.
The normalization approach employed in computing the Invariance Metric $I(\theta)$ is important because in some cases untrained networks collapse all representation together, regardless of translation, object identity, or category. They always output the same class, and their performance is obviously low. But an unnormalized similarity analysis (e.g. using only $T(\theta)$) would return a very high value, giving the impression that the network was indeed performing translation invariance. Such networks would achieve translation invariance through the mechanism of being invariant to anything. 

\section{Invariance Representation Across Horizontal Displacement} \label{appHorizontal}
\begin{figure}[!ht]
\centering
  \includegraphics[width=1\linewidth]{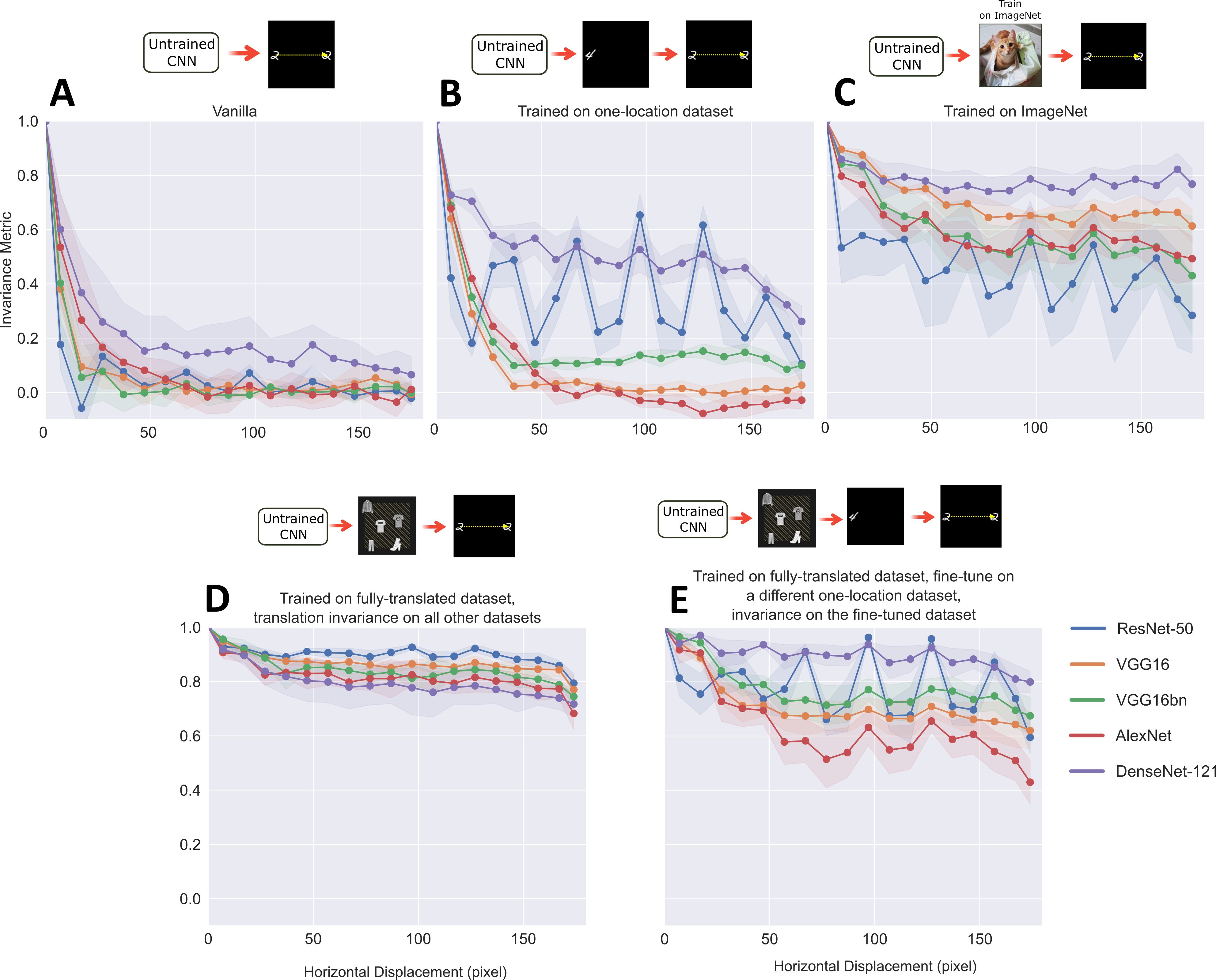}
\caption{Invariance Metric across several horizontal displacement, for different experimental stages. See text for details.}
  \label{FigureAppInvarianceMetricAllStages}
   \end{figure}

In this Section we show in more detail the results of the analysis of invariance representation performed in Section \ref{representation}. The analysis was performed in 5 different stages. We average the results across different data sets. Translation invariance was measured by computing the Invariance Metric (Section \ref{metric}) between a left-most object ($x=25, y=112$) and several horizontal displacement of the same object ($y=[25, 35, ..., 195]$). Results are shown in Figure \ref{FigureAppInvarianceMetricAllStages}.
The different stages were the following:

 \renewcommand{\labelenumi}{\Alph{enumi}.}
 \begin{enumerate}
 \item As a sanity check, we measured translation invariance for each \textit{untrained} network. As expected, this quickly dropped at the smallest displacement.   
 \item We trained each one-location data set, and then computed the Invariance Metric on displacement of the same data set. This was low for all networks but DenseNet-121, and reflects the checkerboard pattern observed in the Density Plots for ResnNet-50.
\item Each network was pretrained on ImageNet. The Invariance Metric was computed on each data set. Performance increase highly. These two latter results reflect accuracy performance in Section \ref{ImageNet}.
\item We pretrained each network on a fully-translated data set. We then computed the invariance metric on displacement of all \textit{other} data sets. This shows that networks pretrained on fully-translated data sets acquired online translation invariance. The effect was even stronger than with ImageNet, probably due to the fact that the translation in the ImageNet data set was a side-effect of a different augmentation process (Random Crop).
\item We pretrained each network on a fully-translated data set, then fine-tuned it on a different one-location data set, and measured its invariance to the fine-tuned data set across several displacement. The difference between the D and E results account for the drop in accuracy shown in Figure \ref{fig:Figure2} (catastrophic interference). This shows that even though networks acquire a strong online translation invariance (D), they may at least partially lose it when fine-tuned (E).  
\end{enumerate}

 \section{Invariance Representation for each data set} \label{appInvarianceAll}
\begin{figure}[!ht]
\centering
  \includegraphics[width=1\linewidth]{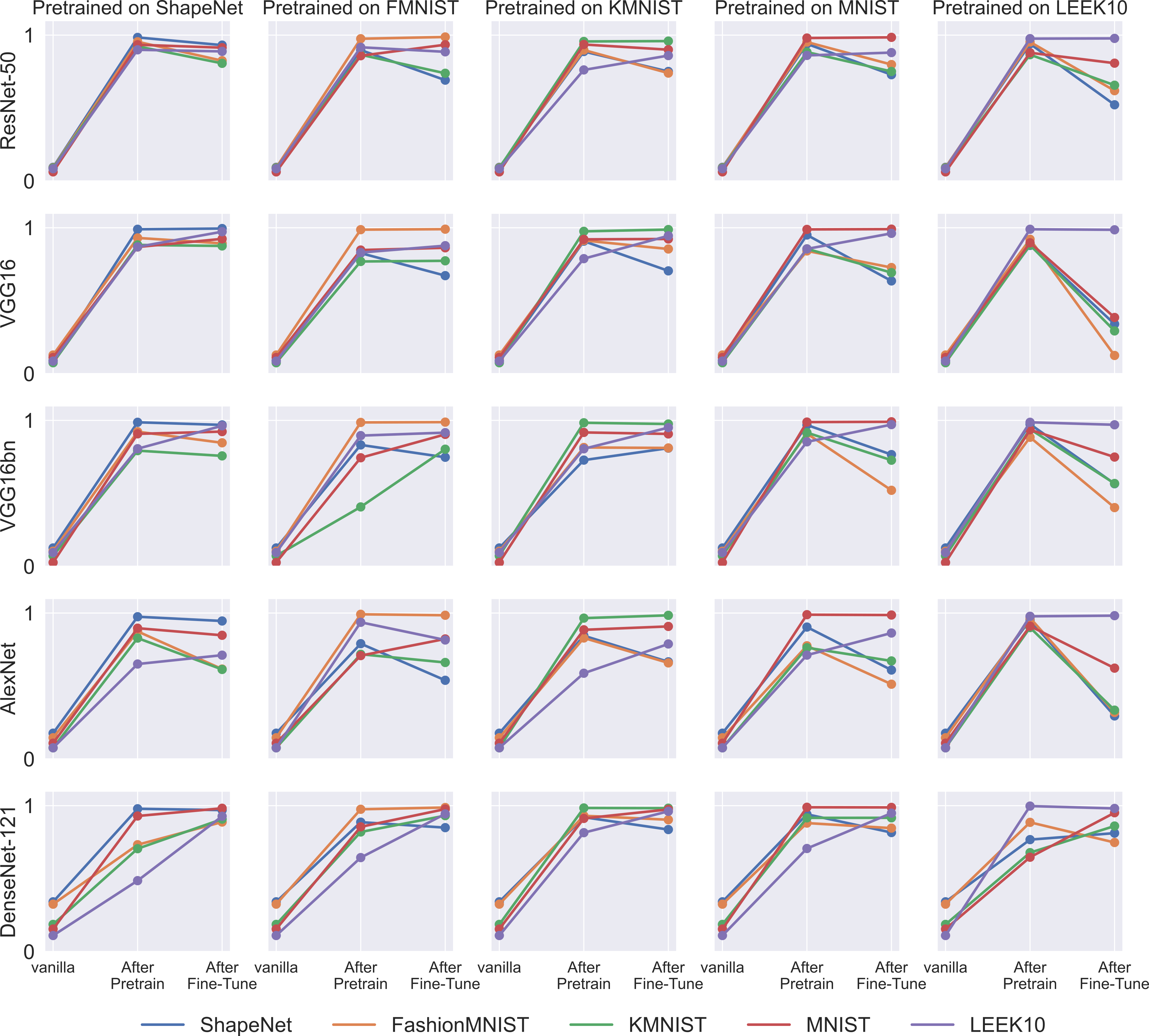}
\caption{Invariance Metric for each network and data set, at three experimental stages. Each network is pretrained on the data set indicated by the column label, and then fine-tuned on the data set indicated by the line color.}
  \label{FigureInvarianceAllNetAlldata set}
   \end{figure}

Figure \ref{FigureInvarianceAllNetAlldata set} shows in more details the result for each data set and each network. We averaged the Invariance Metric across the whole horizontal displacement. Each network is first tested on its untrained version, is then pretrained on the data set indicated by the column position, and then is fine-tuned on the data set indicated by the line color. Each plot show the invariance metric for the fine-tuned data set when untrained, pretrained, and fine-tuned. With this plot is easy to see the effect of fine-tuning on the same data set.
Pretraining on a simple data set reduces the Invariance Metric on more complex data sets. For example, see the drop shared by most networks when pretrained on LEEK10. 

These results are very consistent with the accuracy results shown in Section \ref{Beyond ImageNet} and \ref{AppBeyond}: networks pretrained on complex data set would mostly retain a high degree of cosine similarity after fine-tuning with most of the data sets, meaning that the translation invariance property was preserved. 
   
\section{Training on Limited Translations: Further Tests} \label{AppGeneralizing}
\subsection{Condition 1: Inter-Class Analysis}
\begin{figure}[!ht]
\centering
  \includegraphics[width=1\linewidth]{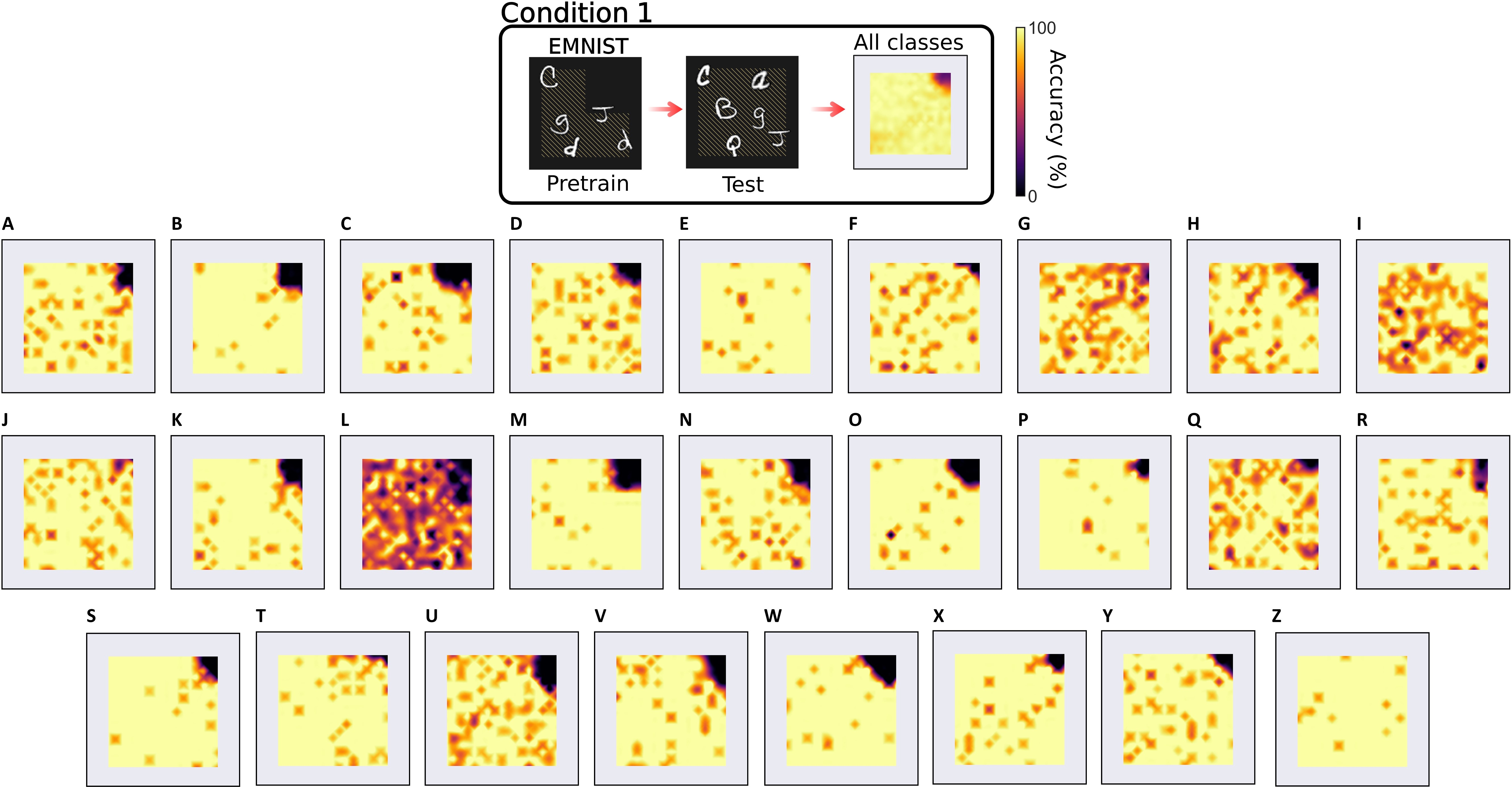}
\caption{Heatmaps for each EMNIST class for the Condition 1 of Experiment 3. The network seem to be perfectly invariant to translation for some classes.}
  \label{FigureAppCond1}
   \end{figure}

We showed that a CNN was only partially able to generalize on untrained locations (Section \ref{Generalizing Translation}). However, we noticed a certain degree of inter-class variability: even though in most cases the results were similar to the average shown in Figure \ref{Figure4}, for a few EMNIST letters the accuracy was high all across the canvas. We show in Figure \ref{FigureAppCond1} the accuracy across the canvas for each  class, noticing that perfect translation invariance was achieved for letter E and Z, and several other classes reached an almost perfect invariance. It is not clear why there was this difference across classes.
Notice that we could find a much lower degree of inter-class variability for Condition 2, and that analysis is therefore not shown here.
\subsection{Condition 1 and 2: Fine-Tuning}
\begin{figure}[!ht]
\centering
  \includegraphics[width=1\linewidth]{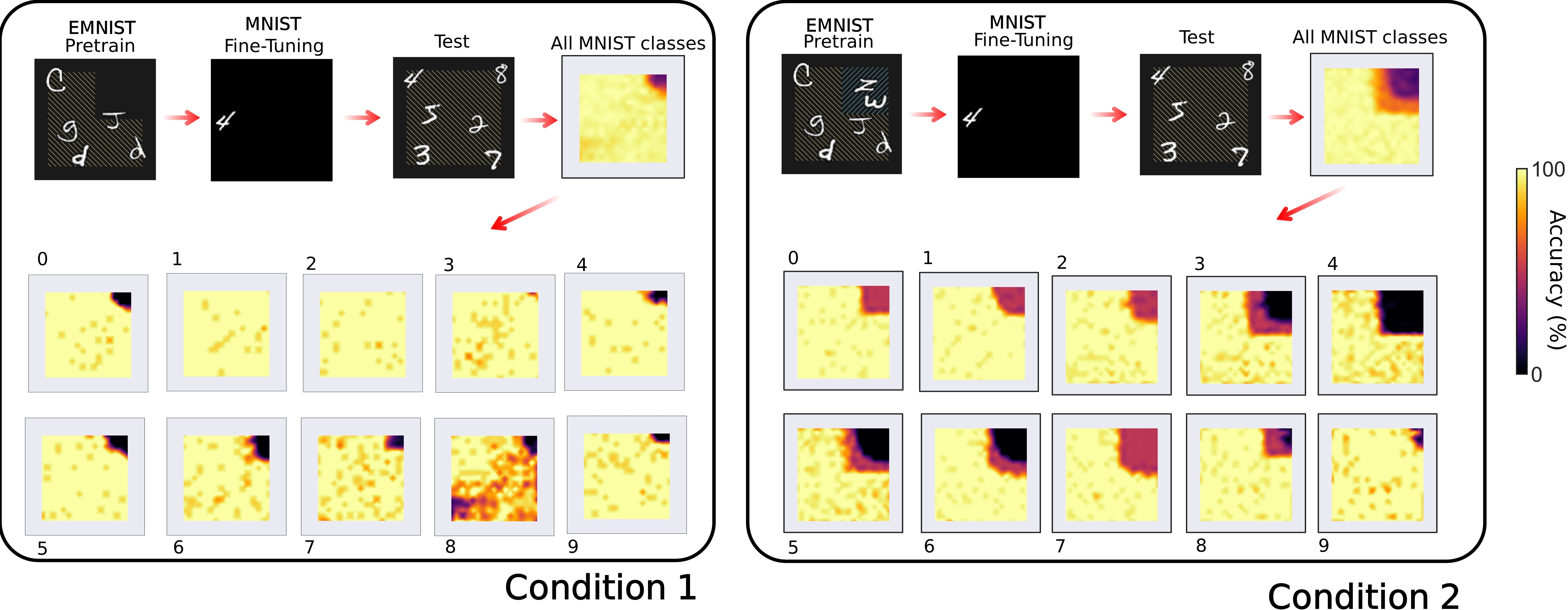}
\caption{Condition 1 and 2 from Experiment 3. Here, instead of testing on the pretrain data set, we fine-tuned on a new data set (MNIST) and tested on the fully-translated version of it. We also show the heatmap for each class, showing a discrete degree of inter-class variability and perfect translation for some classes.}
  \label{FigureAppCond1and2}
   \end{figure}

In Conditions 1 and 2 in Section \ref{Generalizing Translation} we trained a network with EMNIST items by translating them everywhere across the canvas apart from the upper-right quadrant. We then tested the network on the same data set it was trained on (EMNIST), but fully translated. In this section, we show what happens with a experimental design similar to that used in Section \ref{Beyond ImageNet}, that is, by training on a new one-location data set and testing on this fully-translated data set.  
In Condition 1, after having pretrained on the EMNIST data set leaving empty the upper-right quadrant, we fine-tuned the network on one-location MNIST and tested it on a fully-translated MNIST. As shown in Figure \ref{FigureAppCond1and2}, left, we obtained very similar results to those in Figure \ref{FigureAppCond1and2}, that is: the network showed translation invariance on locations all across the pretrained area, but only partial invariance on locations in which it had never been trained. We also analysed the inter-class variability finding that, like in Figure \ref{FigureAppCond1}, the network showed perfect translation invariance for some of the classes in the MNIST data set.

In Condition 2 in Section \ref{Generalizing Translation} we filled the upper-right quadrant with a subset of EMNIST classes, showing that the network learned to selectively recognise classes only on the trained location. In this section, we show what happened when fine-tuning the network on a new one-location data set. This is shown in Figure \ref{FigureAppCond1and2}, right. We can see that the network had retained some kind of bias in that it would incorrectly classify items presented on the upper-right quadrant, but the separation was not as clear as before. As usual, we observed some inter-class variability regarding the network accuracy on the left-out quadrant, but in this case none of the classes reached perfect translation invariance.

\subsection{Condition 3: More Details} \label{Cond3}
\begin{figure}[!ht]
\centering
  \includegraphics[width=0.5\linewidth]{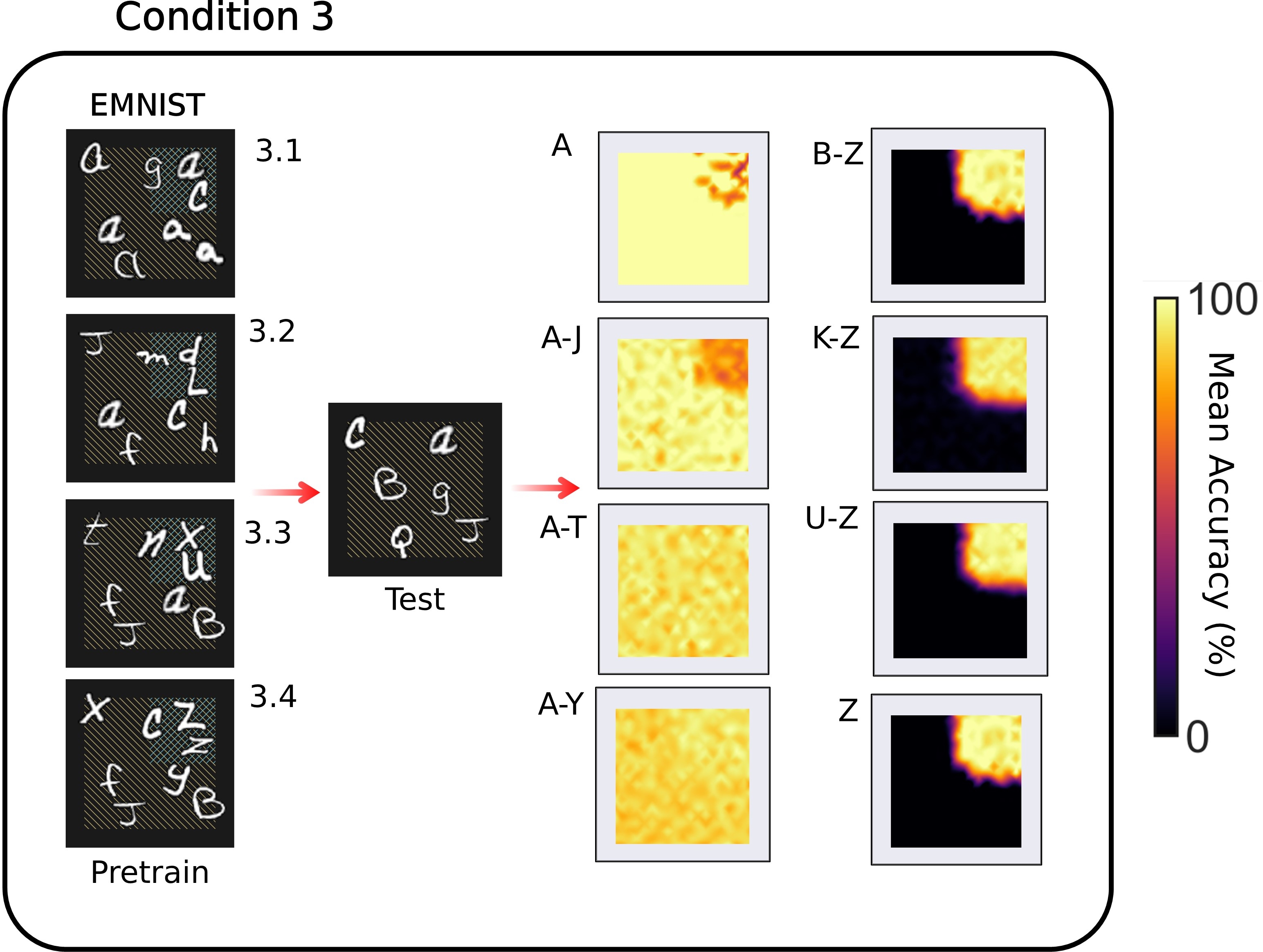}   
\caption{Condition 3 for Experiment 3. This condition was split in 4 sub-conditions, with increasing number of letters trained on the whole canvas. Heatmaps are shown for each group.}
  \label{FigureAppCondition3}
   \end{figure}

In this section we provided more details about the Condition 3 briefly described in Section \ref{Generalizing Translation}. We used a data set structured in this way: a subset of classes from EMNIST was placed anywhere on the canvas (fully-translated); the remaining classes were only placed on the right-upper quadrant. We varied the number of classes trained on the whole canvas in the following way: for Condition 3.1, only the class corresponding to the letter A was trained on the whole canvas; for Condition 3.2, the classes from A to J; for Condition 3.3, the classes from A to T; from Condition 3.4, the classes from A-Y. 
Therefore the amount of classes trained on the whole canvas, for each sub-condition, 1, 10, 20, and 25. 
We then tested the network on the same EMNIST data set, but this time all classes were fully translated. The relevant measure here is the accuracy for the classes presented only on the right-upper quadrant, when then tested on the whole canvas, and in particular whether the network would be able to accurately classify those classes when presented on the area of the canvas in which they were not trained on. The results are shown in Figure \ref{FigureAppCondition3}. We analysed separately the results from the two classes. The interplay between these two groups (fully-translated and partially-translated) is non-trivial, with the Condition 3.1 showing almost perfect translation invariance for the letter A and the condition 3.2 showing a slightly degraded performance for the fully-translated letters A-J on the right-upper quadrant. However, across all sub-conditions, it is clear that the letters trained on the right-upper quadrant were never recognized when presented elsewhere.

\vskip 0.2in
 \bibliography{21-019}

\begin{thebibliography}{52}
\providecommand{\natexlab}[1]{#1}
\providecommand{\url}[1]{\texttt{#1}}
\expandafter\ifx\csname urlstyle\endcsname\relax
  \providecommand{\doi}[1]{doi: #1}\else
  \providecommand{\doi}{doi: \begingroup \urlstyle{rm}\Url}\fi

\bibitem[Azulay and Weiss(2019)]{Azulay2019}
Aharon Azulay and Yair Weiss.
\newblock {Why do deep convolutional networks generalize so poorly to small
  image transformations?}
\newblock \emph{Journal of Machine Learning Research}, 20:\penalty0 1--25,
  2019.
\newblock ISSN 15337928.

\bibitem[Baker et~al.(2018)Baker, Lu, Erlikhman, and Kellman]{Baker2018}
Nicholas Baker, Hongjing Lu, Gennady Erlikhman, and Philip~J. Kellman.
\newblock {Deep convolutional networks do not classify based on global object
  shape}.
\newblock \emph{PLoS Computational Biology}, 14\penalty0 (12):\penalty0 1--43,
  2018.
\newblock ISSN 15537358.
\newblock \doi{10.1371/journal.pcbi.1006613}.

\bibitem[Beaulieu et~al.(2020)Beaulieu, Frati, Miconi, Lehman, Stanley, Clune,
  and Cheney]{Beaulieu2020}
Shawn Beaulieu, Lapo Frati, Thomas Miconi, Joel Lehman, Kenneth~O. Stanley,
  Jeff Clune, and Nick Cheney.
\newblock {Learning to Continually Learn}.
\newblock \emph{arXiv preprint arXiv: 2002.09571}, 2020.
\newblock URL \url{http://arxiv.org/abs/2002.09571}.

\bibitem[Biederman and Cooper(1991)]{BiedermannCooper1991}
I.~Biederman and E.~E. Cooper.
\newblock {Evidence for complete translational and reflectional invariance in
  visual object priming.}
\newblock \emph{Perception}, 20\penalty0 (5):\penalty0 585--593, 6 1991.
\newblock ISSN 03010066.
\newblock \doi{10.1068/p200585}.
\newblock URL \url{https://journals.sagepub.com/doi/10.1068/p200585}.

\bibitem[Blything et~al.(2020)Blything, Biscione, and
  Bowers]{BlythingCommentary2021}
Ryan Blything, Valerio Biscione, and Jeffrey Bowers.
\newblock {A case for robust translation tolerance in humans and CNNs. A
  commentary on Han et al}.
\newblock \emph{arXiv prepring arXiv: 2012.05950}, 12 2020.
\newblock URL \url{http://arxiv.org/abs/2012.05950}.

\bibitem[Blything et~al.(2021)Blything, Biscione, Vankov, Ludwig, and
  Bowers]{Blything2021}
Ryan Blything, Valerio Biscione, Ivan~I Vankov, Casimir J~H Ludwig, and
  Jeffrey~S Bowers.
\newblock {The human visual system and CNNs can both support robust online
  translation tolerance following extreme displacements}.
\newblock \emph{Journal of Vision}, 21\penalty0 (2):\penalty0 1--16, 2021.
\newblock \doi{10.1167/jov.21.2.9}.
\newblock URL \url{https://doi.org/10.1167/jov.21.2.9.}

\bibitem[Bowers et~al.(2016)Bowers, Vankov, and Ludwig]{Bowers2016}
Jeffrey~S. Bowers, Ivan~I. Vankov, and Casimir~J.H. Ludwig.
\newblock {The visual system supports online translation invariance for object
  identification}.
\newblock \emph{Psychonomic Bulletin and Review}, 23\penalty0 (2):\penalty0
  432--438, 4 2016.
\newblock ISSN 15315320.
\newblock \doi{10.3758/s13423-015-0916-2}.
\newblock URL
  \url{https://link.springer.com/article/10.3758/s13423-015-0916-2}.

\bibitem[Chang et~al.(2015)Chang, Funkhouser, Guibas, Hanrahan, Huang, Li,
  Savarese, Savva, Song, Su, Xiao, Yi, and Yu]{Chang2015shapenet}
Angel~X. Chang, Thomas Funkhouser, Leonidas Guibas, Pat Hanrahan, Qixing Huang,
  Zimo Li, Silvio Savarese, Manolis Savva, Shuran Song, Hao Su, Jianxiong Xiao,
  Li~Yi, and Fisher Yu.
\newblock {ShapeNet: An Information-Rich 3D Model Repository}.
\newblock \emph{arXiv prepring arXiv: 1512.03012}, 12 2015.
\newblock URL \url{http://arxiv.org/abs/1512.03012}.

\bibitem[Chen et~al.(2017)Chen, Roig, Isik, Boix, and Poggio]{Chen2017}
Francis~X. Chen, Gemma Roig, Leyla Isik, Xavier Boix, and Tomaso Poggio.
\newblock {Eccentricity dependent deep neural networks: Modeling invariance in
  human vision}.
\newblock \emph{AAAI Spring Symposium - Technical Report}, SS-17-01 -:\penalty0
  541--546, 2017.

\bibitem[Clanuwat et~al.(2018)Clanuwat, Bober-Irizar, Kitamoto, Lamb, Yamamoto,
  and Ha]{KMNIST}
Tarin Clanuwat, Mikel Bober-Irizar, Asanobu Kitamoto, Alex Lamb, Kazuaki
  Yamamoto, and David Ha.
\newblock {Deep Learning for Classical Japanese Literature}.
\newblock \emph{arXiv preprint arXiv: 1812.01718}, 12 2018.
\newblock \doi{10.20676/00000341}.
\newblock URL \url{http://arxiv.org/abs/1812.01718
  http://dx.doi.org/10.20676/00000341}.

\bibitem[Cohen et~al.(2017)Cohen, Afshar, Tapson, and van Schaik]{EMNIST}
Gregory Cohen, Saeed Afshar, Jonathan Tapson, and André van Schaik.
\newblock {EMNIST: an extension of MNIST to handwritten letters}.
\newblock \emph{arXiv preprint arXiv: 1702.05373}, 2 2017.
\newblock URL \url{http://arxiv.org/abs/1702.05373}.

\bibitem[Cooper et~al.(1992)Cooper, Biederman, and
  Hummel]{CooperBiedermann1992}
E.~E. Cooper, I.~Biederman, and J.~E. Hummel.
\newblock {Metric invariance in object recognition: a review and further
  evidence.}
\newblock \emph{Canadian journal of psychology}, 46\penalty0 (2):\penalty0
  191--214, 1992.
\newblock ISSN 00084255.
\newblock \doi{10.1037/h0084317}.
\newblock URL \url{https://pubmed.ncbi.nlm.nih.gov/1451041/}.

\bibitem[Ellis et~al.(1989)Ellis, Allport, Humphreys, and Collis]{Ellis1989}
R.~Ellis, D.~A. Allport, G.~W. Humphreys, and J.~Collis.
\newblock {Varieties of Object Constancy}.
\newblock \emph{The Quarterly Journal of Experimental Psychology Section A},
  41\penalty0 (4):\penalty0 775--796, 11 1989.
\newblock ISSN 14640740.
\newblock \doi{10.1080/14640748908402393}.
\newblock URL \url{/record/1990-11427-001}.

\bibitem[Fiser and Biederman(2001)]{Fiser2001}
József Fiser and Irving Biederman.
\newblock {Invariance of long-term visual priming to scale, reflection,
  translation, and hemisphere}.
\newblock \emph{Vision Research}, 41\penalty0 (2):\penalty0 221--234, 1 2001.
\newblock ISSN 00426989.
\newblock \doi{10.1016/S0042-6989(00)00234-0}.

\bibitem[Fukushima(1980)]{Fukushima1980}
Kunihiko Fukushima.
\newblock {Neocognitron: A self-organizing neural network model for a mechanism
  of pattern recognition unaffected by shift in position}.
\newblock \emph{Biological Cybernetics}, 36\penalty0 (4):\penalty0 193--202, 4
  1980.
\newblock ISSN 03401200.
\newblock \doi{10.1007/BF00344251}.

\bibitem[Funke et~al.(2020)Funke, Borowski, Stosio, Brendel, Wallis, and
  Bethge]{FunkeBorowski2020}
Christina~M Funke, Judy Borowski, Karolina Stosio, Wieland Brendel, Thomas S~A
  Wallis, and Matthias Bethge.
\newblock {The Notorious Difficulty of Comparing Human and Machine Perception}.
\newblock \emph{arXiv preprint arXiv: 2004.09406v1}, 2020.

\bibitem[Furlanello et~al.(2016)Furlanello, Zhao, Saxe, Itti, and
  Tjan]{Furlanello2016}
Tommaso Furlanello, Jiaping Zhao, Andrew~M. Saxe, Laurent Itti, and Bosco~S.
  Tjan.
\newblock {Active Long Term Memory Networks}.
\newblock \emph{arXiv preprint arXiv: 1606.02355}, 6 2016.
\newblock URL \url{http://arxiv.org/abs/1606.02355}.

\bibitem[Furukawa(2017)]{Furukawa2017}
Hidetoshi Furukawa.
\newblock {Deep Learning for Target Classification from SAR Imagery: Data
  Augmentation and Translation Invariance}.
\newblock \emph{arXiv preprint arXiv: 1708.07920}, 8 2017.
\newblock URL \url{http://arxiv.org/abs/1708.07920}.

\bibitem[Geirhos et~al.(2018)Geirhos, Medina~Temme, Rauber, Sch{\"{u}}tt,
  Bethge, Wichmann, Temme, Rauber, Sch{\"{u}}tt, Bethge, and
  Wichmann]{Geirhos2020}
Robert Geirhos, Carlos~R Medina~Temme, Jonas Rauber, Heiko~H Sch{\"{u}}tt,
  Matthias Bethge, Felix~A Wichmann, Carlos R~M Temme, Jonas Rauber, Heiko~H
  Sch{\"{u}}tt, Matthias Bethge, and Felix~A Wichmann.
\newblock {Generalisation in humans and deep neural networks}.
\newblock In S~Bengio, H~Wallach, H~Larochelle, K~Grauman, N~Cesa-Bianchi, and
  R~Garnett, editors, \emph{Advances in Neural Information Processing Systems
  31}, pages 7538--7550. Curran Associates, Inc., 2018.
\newblock URL
  \url{http://papers.nips.cc/paper/7982-generalisation-in-humans-and-deep-neural-networks.pdf
  https://github.com/rgeirhos/generalisation-humans-DNNs.}

\bibitem[Gens and Domingos(2014)]{GensDomingosSymmetry}
Robert Gens and Pedro~M Domingos.
\newblock {Deep Symmetry Networks}.
\newblock In Z~Ghahramani, M~Welling, C~Cortes, N~D Lawrence, and K~Q
  Weinberger, editors, \emph{Advances in Neural Information Processing Systems
  27}, pages 2537--2545. Curran Associates, Inc., 2014.
\newblock URL
  \url{http://papers.nips.cc/paper/5424-deep-symmetry-networks.pdf}.

\bibitem[Girshick et~al.(2014)Girshick, Donahue, Darrell, and
  Malik]{Girshick2014}
Ross Girshick, Jeff Donahue, Trevor Darrell, and Jitendra Malik.
\newblock {Rich feature hierarchies for accurate object detection and semantic
  segmentation}.
\newblock In \emph{Proceedings of the IEEE Computer Society Conference on
  Computer Vision and Pattern Recognition}, pages 580--587. IEEE Computer
  Society, 9 2014.
\newblock ISBN 9781479951178.
\newblock \doi{10.1109/CVPR.2014.81}.

\bibitem[Gong et~al.(2014)Gong, Wang, Guo, and Lazebnik]{Gong2014}
Yunchao Gong, Liwei Wang, Ruiqi Guo, and Svetlana Lazebnik.
\newblock {Multi-scale Orderless Pooling of Deep Convolutional Activation
  Features}.
\newblock \emph{Lecture Notes in Computer Science (including subseries Lecture
  Notes in Artificial Intelligence and Lecture Notes in Bioinformatics)}, 8695
  LNCS\penalty0 (PART 7):\penalty0 392--407, 3 2014.
\newblock URL \url{http://arxiv.org/abs/1403.1840}.

\bibitem[Han et~al.(2019)Han, Yoon, Kwon, Nam, and Kim]{HanYoon2019}
Chihye Han, Wonjun Yoon, Gihyun Kwon, Seungkyu Nam, and Daeshik Kim.
\newblock {Representation of White- and Black-Box Adversarial Examples in Deep
  Neural Networks and Humans: A Functional Magnetic Resonance Imaging Study}.
\newblock \emph{arXiv preprint arXiv: 1905.02422}, 5 2019.
\newblock URL \url{http://arxiv.org/abs/1905.02422}.

\bibitem[Han et~al.(2020)Han, Roig, Geiger, and Poggio]{Han2020b}
Yena Han, Gemma Roig, Gad Geiger, and Tomaso Poggio.
\newblock {Scale and translation-invariance for novel objects in human vision}.
\newblock \emph{Scientific Reports}, 10\penalty0 (1):\penalty0 1--13, 2020.
\newblock ISSN 20452322.
\newblock \doi{10.1038/s41598-019-57261-6}.

\bibitem[He et~al.(2016)He, Zhang, Ren, and Sun]{He2016}
Kaiming He, Xiangyu Zhang, Shaoqing Ren, and Jian Sun.
\newblock {Deep residual learning for image recognition}.
\newblock In \emph{Proceedings of the IEEE Computer Society Conference on
  Computer Vision and Pattern Recognition}, volume 2016-Decem, pages 770--778.
  IEEE Computer Society, 12 2016.
\newblock ISBN 9781467388504.
\newblock \doi{10.1109/CVPR.2016.90}.

\bibitem[Huang et~al.(2017)Huang, Liu, van~der Maaten, and
  Weinberger]{Huang_2017_CVPR}
Gao Huang, Zhuang Liu, Laurens van~der Maaten, and Kilian~Q Weinberger.
\newblock {Densely Connected Convolutional Networks}.
\newblock In \emph{Proceedings of the IEEE Conference on Computer Vision and
  Pattern Recognition (CVPR)}, 7 2017.

\bibitem[Hummel(2013)]{HUmmel2013}
John~E Hummel.
\newblock {Object Recognition}.
\newblock In \emph{Oxford library of psychology. The Oxford handbook of
  cognitive psychology}, number August, pages 1--16. 2013.
\newblock ISBN 9780195376746.
\newblock \doi{10.1093/oxfordhb/9780195376746.013.0003}.
\newblock URL \url{/record/2012-26298-003}.

\bibitem[Javed and White(2019)]{OML2019}
Khurram Javed and Martha White.
\newblock {Meta-Learning Representations for Continual Learning}.
\newblock In H~Wallach, H~Larochelle, A~Beygelzimer,
  F~d{\textbackslash}textquotesingle Alch{\'{e}}-Buc, E~Fox, and R~Garnett,
  editors, \emph{Advances in Neural Information Processing Systems 32}, pages
  1820--1830. Curran Associates, Inc., 2019.
\newblock URL
  \url{http://papers.nips.cc/paper/8458-meta-learning-representations-for-continual-learning.pdf}.

\bibitem[Kauderer-Abrams(2017)]{Kauderer-Abrams2017a}
Eric Kauderer-Abrams.
\newblock {Quantifying Translation-Invariance in Convolutional Neural
  Networks}.
\newblock \emph{arXiv preprint arXiv: 1801.01450v1}, 12 2017.
\newblock URL \url{http://arxiv.org/abs/1801.01450}.

\bibitem[Kayhan and van Gemert(2020)]{SemihKayhan2020}
Osman~S. Kayhan and Jan~C. van Gemert.
\newblock {On Translation Invariance in CNNs: Convolutional Layers Can Exploit
  Absolute Spatial Location}.
\newblock In \emph{2020 IEEE/CVF Conference on Computer Vision and Pattern
  Recognition (CVPR)}, number class 1, pages 14262--14273, 2020.
\newblock \doi{10.1109/cvpr42600.2020.01428}.

\bibitem[Koffka(2013)]{Koffka2013}
K.~Koffka.
\newblock \emph{{Principles of gestalt psychology}}.
\newblock Taylor and Francis, 1 2013.
\newblock ISBN 9781136306815.
\newblock \doi{10.4324/9781315009292}.
\newblock URL \url{https://www.taylorfrancis.com/books/9781315009292}.

\bibitem[Kriegeskorte(2015)]{Kriegeskorte}
Nikolaus Kriegeskorte.
\newblock {Deep Neural Networks: A New Framework for Modeling Biological Vision
  and Brain Information Processing}.
\newblock \emph{Annual Review of Vision Science}, 1\penalty0 (1):\penalty0
  417--446, 11 2015.
\newblock ISSN 2374-4642.
\newblock \doi{10.1146/annurev-vision-082114-035447}.
\newblock URL \url{www.annualreviews.org}.

\bibitem[Krizhevsky et~al.(2012)Krizhevsky, Sutskever, and
  Hinton]{NIPS2012_c399862d}
Alex Krizhevsky, Ilya Sutskever, and Geoffrey~E Hinton.
\newblock {ImageNet Classification with Deep Convolutional Neural Networks}.
\newblock \emph{Advances in Neural Information Processing Systems}, 25, 2012.
\newblock URL \url{http://code.google.com/p/cuda-convnet/
  https://proceedings.neurips.cc/paper/2012/file/c399862d3b9d6b76c8436e924a68c45b-Paper.pdf}.

\bibitem[LeCun(1989)]{LeCun1989}
Yann LeCun.
\newblock {Generalization and network design strategies}.
\newblock \emph{Connectionism in perspective}, 19:\penalty0 143--155, 1989.
\newblock URL
  \url{https://nyuscholars.nyu.edu/en/publications/generalization-and-network-design-strategies}.

\bibitem[LeCun and Bengio(1995)]{LeCun1995}
Yann LeCun and Yoshua Bengio.
\newblock {Convolutional Networks for Images, Speech, and Time-Series}.
\newblock \emph{The handbook of brain theory and neural networks}, 3361, 1995.

\bibitem[LeCun et~al.(1990)LeCun, Boser, Denker, Henderson, Howard, Hubbard,
  Jackel, Cun, Henderson, Le~Cun, Denker, Henderson, Howard, Hubbard, and
  Jackel]{LeCun1990}
Yann LeCun, Bernhard~E Boser, John~S Denker, Donnie Henderson, R~E Howard,
  Wayne~E Hubbard, Lawrence~D Jackel, Le~Cun, Jackel Henderson, Y~Le~Cun,
  John~S Denker, Donnie Henderson, R~E Howard, Wayne~E Hubbard, and Lawrence~D
  Jackel.
\newblock {Handwritten Digit Recognition with a Back-Propagation Network}.
\newblock In D~S Touretzky, editor, \emph{Advances in Neural Information
  Processing Systems 2}, pages 396--404. Morgan-Kaufmann, 1990.
\newblock URL
  \url{http://papers.nips.cc/paper/293-handwritten-digit-recognition-with-a-back-propagation-network.pdf}.

\bibitem[LeCun et~al.(1998)LeCun, Bottou, Bengio, and Haffner]{LeCun1998}
Yann LeCun, Léon Bottou, Yoshua Bengio, and Patrick Haffner.
\newblock {Gradient-based learning applied to document recognition}.
\newblock \emph{Proceedings of the IEEE}, 86\penalty0 (11):\penalty0
  2278--2323, 1998.
\newblock ISSN 00189219.
\newblock \doi{10.1109/5.726791}.

\bibitem[Leek et~al.(2016)Leek, Roberts, Oliver, Cristino, and Pegna]{Leek2016}
E.~Charles Leek, Mark Roberts, Zoe~J. Oliver, Filipe Cristino, and Alan~J.
  Pegna.
\newblock {Early differential sensitivity of evoked-potentials to local and
  global shape during the perception of three-dimensional objects}.
\newblock \emph{Neuropsychologia}, 89:\penalty0 495--509, 8 2016.
\newblock ISSN 18733514.
\newblock \doi{10.1016/j.neuropsychologia.2016.07.006}.

\bibitem[Lenc and Vedaldi(2019)]{Lenc2019}
Karel Lenc and Andrea Vedaldi.
\newblock {Understanding Image Representations by Measuring Their Equivariance
  and Equivalence}.
\newblock \emph{International Journal of Computer Vision}, 127\penalty0
  (5):\penalty0 456--476, 5 2019.
\newblock ISSN 15731405.
\newblock \doi{10.1007/s11263-018-1098-y}.
\newblock URL \url{https://doi.org/10.1007/s11263-018-1098-y}.

\bibitem[Li and Hoiem(2016)]{Li2016}
Zhizhong Li and Derek Hoiem.
\newblock {Learning without forgetting}.
\newblock In \emph{Lecture Notes in Computer Science (including subseries
  Lecture Notes in Artificial Intelligence and Lecture Notes in
  Bioinformatics)}, volume 9908 LNCS, pages 614--629. Springer Verlag, 2016.
\newblock ISBN 9783319464923.
\newblock \doi{10.1007/978-3-319-46493-0{\_}37}.
\newblock URL
  \url{https://link.springer.com/chapter/10.1007/978-3-319-46493-0_37}.

\bibitem[Ma and Peters(2020)]{Ma2020}
Wei~Ji Ma and Benjamin Peters.
\newblock {A neural network walks into a lab: towards using deep nets as models
  for human behavior}.
\newblock \emph{arXiv preprint arXiv: 2005.02181}, pages 1--39, 2020.
\newblock URL \url{http://arxiv.org/abs/2005.02181}.

\bibitem[Marcos et~al.(2016)Marcos, Volpi, and Tuia]{Marcos2016}
Diego Marcos, Michele Volpi, and Devis Tuia.
\newblock {Learning rotation invariant convolutional filters for texture
  classification}.
\newblock In \emph{Proceedings - International Conference on Pattern
  Recognition}, volume~0, pages 2012--2017. Institute of Electrical and
  Electronics Engineers Inc., 1 2016.
\newblock ISBN 9781509048472.
\newblock \doi{10.1109/ICPR.2016.7899932}.
\newblock URL \url{http://arxiv.org/abs/1604.06720
  http://dx.doi.org/10.1109/ICPR.2016.7899932 http://www.outex.oulu.fi}.

\bibitem[Marcus(2018)]{DeepLearningMarcus2018}
Gary Marcus.
\newblock {Deep Learning: A Critical Appraisal}.
\newblock \emph{arXiv preprint arXiv: 1801.00631}, 1 2018.
\newblock URL \url{http://arxiv.org/abs/1801.00631}.

\bibitem[McCloskey and Cohen(1989)]{McCloskey1989}
Michael McCloskey and Neal~J. Cohen.
\newblock {Catastrophic Interference in Connectionist Networks: The Sequential
  Learning Problem}.
\newblock \emph{Psychology of Learning and Motivation - Advances in Research
  and Theory}, 24\penalty0 (C):\penalty0 109--165, 1 1989.
\newblock ISSN 00797421.
\newblock \doi{10.1016/S0079-7421(08)60536-8}.

\bibitem[Montero et~al.(2021)Montero, Ludwig, Ponte~Costa, Malhotra, and
  Bowers]{Montero2021TheGeneralisation}
Milton~L Montero, Casimir J~H Ludwig, Rui Ponte~Costa, Gaurav Malhotra, and
  Jeffrey~S Bowers.
\newblock {The Role of Disentanglement in Generalisation}.
\newblock In \emph{In International Conference on Learning Representations},
  2021.

\bibitem[Richards et~al.(2019)Richards, Lillicrap, Beaudoin, Bengio, Bogacz,
  Christensen, Clopath, Costa, de~Berker, Ganguli, Gillon, Hafner, Kepecs,
  Kriegeskorte, Latham, Lindsay, Miller, Naud, Pack, Poirazi, Roelfsema,
  Sacramento, Saxe, Scellier, Schapiro, Senn, Wayne, Yamins, Zenke, Zylberberg,
  Therien, and Kording]{Richards2019}
Blake~A. Richards, Timothy~P. Lillicrap, Philippe Beaudoin, Yoshua Bengio,
  Rafal Bogacz, Amelia Christensen, Claudia Clopath, Rui~Ponte Costa, Archy
  de~Berker, Surya Ganguli, Colleen~J. Gillon, Danijar Hafner, Adam Kepecs,
  Nikolaus Kriegeskorte, Peter Latham, Grace~W. Lindsay, Kenneth~D. Miller,
  Richard Naud, Christopher~C. Pack, Panayiota Poirazi, Pieter Roelfsema, João
  Sacramento, Andrew Saxe, Benjamin Scellier, Anna~C. Schapiro, Walter Senn,
  Greg Wayne, Daniel Yamins, Friedemann Zenke, Joel Zylberberg, Denis Therien,
  and Konrad~P. Kording.
\newblock {A deep learning framework for neuroscience}.
\newblock \emph{Nature Neuroscience}, 22\penalty0 (11):\penalty0 1761--1770,
  2019.
\newblock ISSN 15461726.
\newblock \doi{10.1038/s41593-019-0520-2}.
\newblock URL \url{http://dx.doi.org/10.1038/s41593-019-0520-2}.

\bibitem[Simonyan and Zisserman(2014)]{VGG16}
Karen Simonyan and Andrew Zisserman.
\newblock {Very deep convolutional networks for large-scale image recognition}.
\newblock \emph{arXiv preprint arXiv: 1409.1556}, 2014.
\newblock URL \url{http://www.robots.ox.ac.uk/}.

\bibitem[Srivastava et~al.(2019)Srivastava, Ben-Yosef, and
  Boix]{Srivastava2019}
Sanjana Srivastava, Guy Ben-Yosef, and Xavier Boix.
\newblock {Minimal Images in Deep Neural Networks: Fragile Object Recognition
  in Natural Images}.
\newblock \emph{7th International Conference on Learning Representations, ICLR
  2019}, 2 2019.
\newblock URL \url{http://arxiv.org/abs/1902.03227}.

\bibitem[Vankov and Bowers(2020)]{VankovBowers2020}
Ivan~I. Vankov and Jeffrey~S. Bowers.
\newblock {Trainin Neural Networks to encode symbols enables combinatorial
  generlization}.
\newblock \emph{Philosophical Transactions of the Royal Society B: Biological
  Sciences}, 375\penalty0 (1791):\penalty0 20190309, 2 2020.
\newblock ISSN 0962-8436.
\newblock \doi{10.1098/rstb.2019.0309}.
\newblock URL
  \url{https://royalsocietypublishing.org/doi/10.1098/rstb.2019.0309}.

\bibitem[Xiao et~al.(2017)Xiao, Rasul, and Vollgraf]{fashionMNIST}
Han Xiao, Kashif Rasul, and Roland Vollgraf.
\newblock {Fashion-MNIST: a Novel Image Dataset for Benchmarking Machine
  Learning Algorithms}.
\newblock \emph{arXiv preprint arXiv: 1708.07747}, 8 2017.
\newblock URL \url{http://arxiv.org/abs/1708.07747}.

\bibitem[Zhang(2019)]{Zhang2019}
Richard Zhang.
\newblock {Making Convolutional Networks Shift-Invariant Again}.
\newblock \emph{36th International Conference on Machine Learning, ICML 2019},
  2019-June:\penalty0 12712--12722, 4 2019.
\newblock URL \url{http://arxiv.org/abs/1904.11486}.

\bibitem[Zhuang et~al.(2020)Zhuang, Yan, Nayebi, Schrimpf, Frank, DiCarlo, and
  Yamins]{Zhuang2020}
Chengxu Zhuang, Siming Yan, Aran Nayebi, Martin Schrimpf, Michael~C. Frank,
  James~J. DiCarlo, and Daniel L.~K. Yamins.
\newblock {Unsupervised Neural Network Models of the Ventral Visual Stream}.
\newblock \emph{bioRxiv}, page 2020.06.16.155556, 2020.
\newblock \doi{10.1101/2020.06.16.155556}.
\newblock URL
  \url{https://www.biorxiv.org/content/10.1101/2020.06.16.155556v1%0Ahttps://www.biorxiv.org/content/10.1101/2020.06.16.155556v1.abstract}.

\end{thebibliography}
\end{document}